\documentclass{article}

\usepackage{PRIMEarxiv}
\usepackage{graphicx}
\usepackage{natbib}
\usepackage{booktabs}
\usepackage{multirow}
\usepackage{makecell}
\usepackage{tabularx}
\usepackage{hyperref}

\title{Robustness Enhancement in Neural Networks with Alpha-Stable Training Noise
}

\author{
  Xueqiong Yuan\footnotemark[1], Jipeng Li\footnotemark[1], Ercan Engin Kuruoğlu\footnotemark[2] \\
  Tsinghua-Berkeley Shenzhen Institute \\
  Tsinghua University \\
  \texttt{\{xq-yuan22, lijipeng22\}@mails.tsinghua.edu.cn, kuruoglu@sz.tsinghua.edu.cn} \\
}

\begin{document}
\maketitle

\renewcommand*{\thefootnote}{\fnsymbol{footnote}}
\footnotetext[1]{These authors contributed equally.}
\renewcommand*{\thefootnote}{\arabic{footnote}}
\begin{abstract}
 With the increasing use of deep learning on data collected by non-perfect sensors and in non-perfect environments, the robustness of deep learning systems has become an important issue. A common approach for obtaining robustness to noise has been to train deep learning systems with data augmented with Gaussian noise. In this work, we challenge the common choice of Gaussian noise and explore the possibility of stronger robustness for non-Gaussian impulsive noise, specifically $\alpha$-stable noise. Justified by the Generalized Central Limit Theorem and evidenced by observations in various application areas, $\alpha$-stable noise is widely present in nature. By comparing the testing accuracy of models trained with Gaussian noise and $\alpha$-stable noise on data corrupted by different noise, we find that training with $\alpha$-stable noise is more effective than Gaussian noise, especially when the dataset is corrupted by impulsive noise, thus improving the robustness of the model. The generality of this conclusion is validated through experiments conducted on various deep learning models with image and time series datasets, and other benchmark corrupted datasets. Consequently, we propose a novel data augmentation method that replaces Gaussian noise, which is typically added to the training data, with $\alpha$-stable noise.
\end{abstract}

\section{INTRODUCTION}

Deep learning models have become an effective tool for solving various tasks, including pattern recognition, semantic segmentation, natural language processing, etc. The performance of deep learning models is related to the quality of the input data. It is common for real-life data to be corrupted, both naturally (e.g., noise) and maliciously (e.g., attacks). Therefore, robustness is an important criterion for evaluating machine learning models, which refers to the ability of a model to maintain stable and accurate performance even in the presence of corrupted data. It goes beyond achieving reliable performance on clean datasets and encompasses the resilience of models against noise, distortions, variation, and adversarial attacks. Robustness not only ensures the generalization and reliability of the model, but also makes the model more interpretable.

One of the frequently occurring perturbations is additive random noise, which is the focus of our work. Gaussian noise is a commonly observed type of noise and is often present in various systems, including electronic components. The Central Limit Theorem (CLT) suggests that under certain conditions, the sum or average of a large number of independent and identically distributed (i.i.d.) random variables will approximately follow a Gaussian distribution, regardless of the original distribution of the variables. Furthermore, the Gaussian distribution is defined with only two parameters, mean and variance, given which it is the maximum entropy distribution. Due to these excellent mathematical properties, the noise is generally assumed to be Gaussian. In order to enhance robustness in the presence of noise, data augmentation methods have been proposed that involve adding Gaussian noise to the training data. This technique is commonly referred to as noise injection and has been theoretically demonstrated to be equivalent to adding penalty terms to the cost function \citep{reed1992regularization,DBLP:journals/tsmc/Matsuoka92,bishop1995training, DBLP:journals/neco/An96, DBLP:journals/neco/GrandvaletCB97}, thus improving the model generalization ability.

\begin{figure}[h]
\centering
\includegraphics{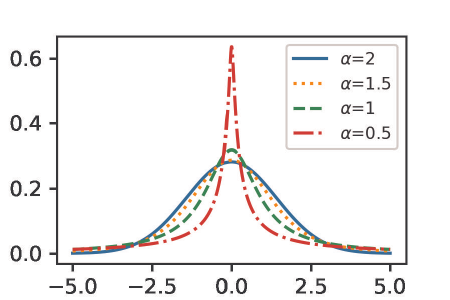}
\caption{Probability density function curves of symmetric $\alpha$-stable distribution with $\alpha=2,1.5,1,0.5$}
\label{pdf}
\end{figure}

However, in addition to Gaussian noise, it is possible that the environmental noise is non-Gaussian and impulsive. In this case, signals and noise are prone to displaying abrupt peaks or occasional bursts more frequently than Gaussian distribution. For example, images are often corrupted by impulsive noise due to noisy sensors or transmission channels \citep{windyga2001fast,karakucs2018generalized,zhan2019recovery,DBLP:journals/sivp/KarakusKA20}. Radar and sonar data \citep{7472965}, as well as medical data such as ECG \citep{Singh_Bhole_Sharma_2017}, MRI \citep{Kalavathi_Priya_2016}, OCT \citep{lee2023deep}, etc., are also frequently subjected to interference from impulsive noise. The $\alpha$-stable distribution has been suggested to model impulsive noise \citep{Mandelbrot1963, DBLP:journals/pieee/ShaoN93}. The $\alpha$-stable distribution is a generalization of Gaussian distribution, of which Gaussian distribution is a special case. The $\alpha$-stable distribution allows flexible characterization of impulsive noise with a characteristic exponent $\alpha \in (0, 2]$ to control the heavy-tailedness (impulsiveness). The smaller the $\alpha$, the heavier the tail, and the more impulsive the noise. Figure \ref{pdf} shows the probability density function curve of $\alpha$-stable distribution with $\alpha$ taking different values.

The theoretical justification for choosing the $\alpha$-stable distribution to model noise is Generalized Central Limit Theorem (GCLT). It is known that the CLT requires that the random variables have finite variance. However, the GCLT is more powerful than the CLT, stating that the sum of infinite i.i.d. random variables, with or without finite variance, converges to the $\alpha$-stable distribution \citep{levy1937theorie}.

In practice, the $\alpha$-stable distribution has been used to model switching noise on telephone lines \citep{stuck1974statistical}, laser gyroscope noise \citep{shen2015observation}, gradient noise produced in backpropagation \citep{DBLP:journals/corr/abs-1912-00018}, noise in powerline communication systems \citep{DBLP:journals/sivp/KarakusKA20}, etc. The $\alpha$-stable distribution has been adopted in several areas of machine learning, such as adversarial attacks \citep{DBLP:conf/eusipco/SrinivasanKMSN19} and heavy-tailed behavior in stochastic gradient descent \citep{DBLP:journals/corr/abs-1912-00018}. However, the $\alpha$-stable noise has failed to receive sufficient attention in the field of robustness of deep learning models. The examination of whether deep learning models remain robust when faced with data corrupted by $\alpha$-stable noise, as well as the methods for robustness enhancement in the presence of $\alpha$-stable noise, are meaningful topics worth exploring but currently neglected. If we train deep learning systems on data with Gaussian noise and test on $\alpha$-stable noisy data, will the model still have good robustness? Can replacing the Gaussian noise in training data with $\alpha$-stable noise improve the model robustness? These are the main questions we would like to answer with our study.

In this paper, we question the conventional use of Gaussian noise and investigate the potential for enhanced robustness with non-Gaussian impulsive noise, modelled by $\alpha$-stable distribution. To be precise, we replace Gaussian training noise with $\alpha$-stable one in the data augmentation phase and test the model robustness. Our key contributions can be summarized as follows: (\romannumeral1) We conduct a comprehensive experimental evaluation using various datasets, including MNIST, CIFAR10, ECG200 and LIBRAS, to assess the performance of deep learning models, including Fully Connected Neural Networks (FCNs), Visual Geometry Groups (VGGs), Residual Networks (ResNets) and Long Short-Term Memory Networks (LSTMs), trained with different types of $\alpha$-stable noise. (\romannumeral2) By comparing the accuracy and robustness of models trained with Gaussian noise and $\alpha$-stable noise on both image and time series data, we provide insights into the effectiveness of $\alpha$-stable noise for improving model robustness. It is found that when using multiple noise with different $\alpha$ values, or Cauchy noise ($\alpha=1$) for data augmentation, it maintains relatively high accuracy across test sets with varying degrees of corruption, demonstrating its universality. (\romannumeral3) We conduct experiments on benchmark corrupted datasets, MNIST-C and CIFAR10-C, encompassing various types of corruptions to evaluate our $\alpha$-stable noise trained models. Our findings indicate that models trained with $\alpha$-stable noise exhibit sparser parameters and demonstrate superior generalizability compared to models trained with Gaussian noise. Therefore, we propose a new data augmentation method to replace Gaussian noise with $\alpha$-stable noise. Overall, our findings show that $\alpha$-stable noise is effective for improving neural network robustness, contributing to the development of more robust and reliable deep learning models.

\section{RELATED WORK}\label{sec:related work}

\subsection{Robustness of Neural Networks}

There have been many empirical and theoretical analyses about the robustness of neural networks in the face of various types of perturbations, including structured transformation \citep{DBLP:conf/icml/LarochelleECBB07}, adversarial perturbations \citep{DBLP:journals/corr/SzegedyZSBEGF13, DBLP:conf/nips/FawziMF16, DBLP:journals/ml/FawziFF18}, random noise 
\citep{DBLP:conf/nips/FawziMF16, DBLP:journals/ml/FawziFF18} and universal perturbations \citep{DBLP:conf/cvpr/Moosavi-Dezfooli17}. They test the models on perturbed data, and observe the prediction performance. Neural networks have different robustness to random noise, adversarial perturbations and structural transformations. Universal perturbations are observed to have both cross-image and cross-model universality.

It is worth clarifying that our work is concerned with random noise, which differs from adversarial perturbation in that random noise is not intentionally designed, but naturally generated during the data generation or collection, whereas adversarial perturbation is carefully designed, usually through optimization methods to find the minimum perturbation that makes the model make wrong predictions.

\subsection{Noise Injection}

A frequently used method to improve the robustness of neural networks is data augmentation, a technique of artificially expanding a dataset by applying various transformations or modifications to the existing data samples. One branch of data augmentation is noise injection, which means introducing random noise in training process. Early experimental studies have found that training with noisy signals can improve generalization performance \citep{plaut1986experiments, sietsma1988neural}. It is proved that the method of introducing additive noise is asymptotically consistent \citep{holmstrom1992using}. Moreover, several mathematical analyses have been performed based on Taylor expansion of the cost function, theoretically showing that under certain assumptions, training with random input noise is equivalent to Lagrangian regularization with a derivative regularizer \citep{reed1992regularization}, a sensitivity penalty term to the loss function \citep{DBLP:journals/tsmc/Matsuoka92}, or Tikhonov regularization \citep{bishop1995training}, thus improving the model generalibility. \citep{DBLP:journals/neco/An96} challenges the previous three studies and argues that adding noise is not exactly equivalent to the regularization method. This is because noise not only introduces a regularization term to the cost function, but also a term depending on the fitting residues, which is not noticed by the aforementioned studies. \citep{DBLP:journals/neco/GrandvaletCB97} connect Gaussian noise injection and heat kernel, providing a new insight to explain the effect of noise injection on the cost function. In order to prevent the adverse effect of noise injection on the cost function, a modified cost function is proposed \citep{DBLP:conf/nnsp/SeghouaneMF02}. It is also shown that noise benefits accelerating back propagation \citep{DBLP:conf/ijcnn/AudhkhasiOK13, DBLP:journals/nn/KoskoAO20}.

 In addition to the above theoretical research, there are also plenty of empirical studies on the effectiveness of training with noise. Noise injection is found to be as effective as or even outperform weight decay and early stopping \citep{zur2009noise}. It has been also revealed that by strategically incorporating carefully tuned additive noise patterns during training on clean samples, we can achieve superior performance compared to most existing state-of-the-art defense methods against common corruptions \citep{DBLP:journals/corr/abs-2001-06057, Rusak_2020}. The effectiveness of noise injection in improving generalization performance has been demonstrated experimentally in various scenarios of deep learning, including discrete time backpropagation trained network \citep{DBLP:journals/ijon/ReyesD01}, ensemble learning \citep{DBLP:journals/isci/Zhang07,ahn2020neural}, time series forecasting \citep{4738222}, ECG signal classification \citep{DBLP:journals/informaticaSI/Ochoa-BrustMFGM19,venton2021robustness}, speech recognition \citep{DBLP:journals/ejasmp/YinLZLWTZL15}, inverse problems \citep{isaev2016training,DBLP:conf/icann/IsaevBDLVD18}, and decentralized training \citep{DBLP:conf/pkdd/AdilovaPS18}. It is also discovered that noise level has an effect on generalization improvement \citep{DBLP:journals/ijon/ReyesD01,4738222}, and methods of selecting noise are proposed \citep{holmstrom1992using,DBLP:conf/ssci/Moreno-BareaSJU18,ning2021improving}.
 
 \begin{table}[b]
\caption{Dataset Information}
\label{dataset}
    \begin{center}
		\begin{small}
			\begin{tabular}{ccccc}
				\toprule
				Dataset & \makecell[c]{Shape of\\Samples} & \makecell[c]{Training \\Set Size} & \makecell[c]{Testing \\Set Size} & \makecell[c]{\# of\\Classes} \\
				\midrule
				MNIST & (28,28,1) & 60000 & 10000 & 10\\
                CIFAR10 & (32,32,3) & 50000 & 10000 & 10\\
                ECG200 & (96,1) & 100 & 100 & 2\\
                LIBRAS & (45,2) & 90 & 90 & 15\\
				\bottomrule& 
			\end{tabular}
		\end{small}
	\end{center}
\end{table}

The majority of existing research has focused on Gaussian noise, which is a special case of the $\alpha$-stable distribution where $\alpha$ is 2. There have also been studies that utilize speckle noise for data augmentation \citep{Rusak_2020}, but in fact, speckle noise is also a form of heavy-tailed noise. To the best of our knowledge, this is the first study to investigate the effectiveness of using $\alpha$-stable noise for data augmentation.

 \section{METHODOLOGY} \label{sec:methodology}
\subsection{Datasets}

In the study, different $\alpha$-stable noises are added to classical datasets, including two image datasets, MNIST \citep{lecun-mnisthandwrittendigit-2010} and CIFAR10 \citep{Krizhevsky09learningmultiple}, and two time series datasets, ECG200 \citep{10.5555/935627} and LIBRAS \citep{Dua:2019}. MNIST is a dataset of hand-written digit images with 10 classes (digits 0 to 9) and 70,000 images. CIFAR10 is a dataset of 60,000 color images with 10 different categories. ECG200 is a dataset of electrocardiograms, each series capturing the electrical activity during one heartbeat. The dataset consists of two categories: normal heartbeat and myocardial infarction. LIBRAS, which stands for "Lingua BRAsileira de Sinais", is the official Brazilian sign language. It contains 15 different types of hand movements. The hand movements are represented as bi-dimensional curves that trace the path of the hand over time. Different types of datasets are used to demonstrate the generality of our approach. All of these datasets are related to classification tasks. The shape of samples and the size of the training set and testing set of all datasets are shown in Table \ref{dataset}.

In addition to the above datasets, we also select benchmark datasets on the neural network robustness to common corruptions and perturbations, including MNIST-C\citep{mu2019mnist} and CIFAR10-C\citep{hendrycks2019benchmarking}, to evaluate our proposed data augmentation method. These datasets are the corrupted versions of MNIST and CIFAR10, respectively, and they encompass a wide range of common perturbations, including Gaussian noise, impulse noise, blur, weather conditions, etc. These two datasets are used solely for testing purposes and are not used for training.

\subsection{Symmetric $\alpha$-Stable Noise}
Our noise is generated from symmetric $\alpha$-stable distribution \citep{samorodnitsky1994stable}, which is defined by the Fourier Transform of its characteristic function,
\begin{equation}
    f(x;\alpha,\gamma,\delta)=\frac{1}{2\pi}\int_{-\infty}^{+\infty}\phi(t)e^{-itx}dt,
\end{equation}
where $\phi(t)$ can be expressed as
\begin{equation}
    \phi(t)=\exp\left\{it\delta-\gamma|t|^{\alpha}\right\}.
\end{equation}
Here $\delta$ is the location parameter, $\gamma$ is the dispersion parameter, and $\alpha$ is the characteristic exponent, which controls the thickness of the tails of the distribution and ranges from 0 to 2. Smaller $\alpha$ means heavier tails. When $\alpha=2$, it degenerates to Gaussian distribution. When $\alpha=1$, it degenerates to Cauchy distribution. When $\alpha<2$, the variance is infinite and when $\alpha<1$, the finite mean does not exist. We control the $\alpha$ of the noise to explore its influence on the model, as will be described in detail in the next section.

\section{EXPERIMENTS} \label{sec:exp}

\subsection{Noisy Dataset Generation} 

\begin{table}[t]
\caption{Dispersion Parameter}
\label{sigma}
	\begin{center}
		\begin{small}
			\begin{tabular}{cc}
				\toprule
				Dataset & $\gamma$ \\
				\midrule
				MNIST & from 0.035 to 0.35 \\
                CIFAR10 & from 0.021 to 0.35 \\
                ECG200 & from 0.007 to 0.35 \\
                LIBRAS & from 0.007 to 0.21 \\
				\bottomrule& 
			\end{tabular}
		\end{small}
	\end{center}
\end{table}

To generate $\alpha$-stable noise, we utilize the ''levy\_stable'' module from the scipy.stats package in Python.  

Different levels of $\alpha$-stable noise are introduced to each image or time series, with the level of noise controlled by varying the value of $\alpha$. We set the location parameter $\delta$ to 0. The tails of the distribution become thicker as the $\alpha$ value decreases, but it is important to note that the severity of noise is also influenced by the value of dispersion parameter $\gamma$. Therefore, we select a range of different $\gamma$ values for each $\alpha$ and conduct several experiments to find the corresponding optimal $\gamma$ values. The range of $\gamma$ choices is illustrated in Table \ref{sigma}.

To create the single-noise dataset, we replicate the dataset ten times \citep{zantedeschi2017efficient} and apply i.i.d. $\alpha$-stable noise to all images and time series. The values of $\alpha$ used are 2, 1.9, 1.5, 1.3, 1, 0.9, and 0.5. In the training set, we also include the clean dataset. For image data, we first scale it to the range of 0 to 1. After adding noise, the values of each pixel is clipped to ensure they remain within the range of 0 to 1. In our experiments, we conduct additional trials using smaller values for the parameter $\alpha$. However, we observe that when $\alpha$ is set to a value smaller than 0.5, approximately one third of the pixel values in the image dataset become clipped. As a result, the reliability of the obtained results is compromised. Therefore, we determine that the range of $\alpha$ values should start from a minimum value of 0.5. The time series data are not clipped.

In the combined-noise scenario, we utilize two distinct methods: multiple noise and mixture noise. We denote the number of samples in the dataset as $N$. To construct the multiple noise dataset, we duplicate the original dataset twice and apply i.i.d. $\alpha$-stable noise to these $2 \times N$ images or time series. This process is repeated for $\alpha$ values of 2, 1.9, 1.5, 1.3, 1, and 0.9 (Since the inclusion of 0.5 does not result in better performance, it is excluded from the combined-noise scenario). Additionally, we retain the original clean dataset without any noise, resulting in a dataset that combines samples with different $\alpha$ values and clean data. Consequently, the total number of samples in this dataset is $(2 \times 6+1) \times N$. 

\begin{figure}[h]
\begin{center}
\includegraphics{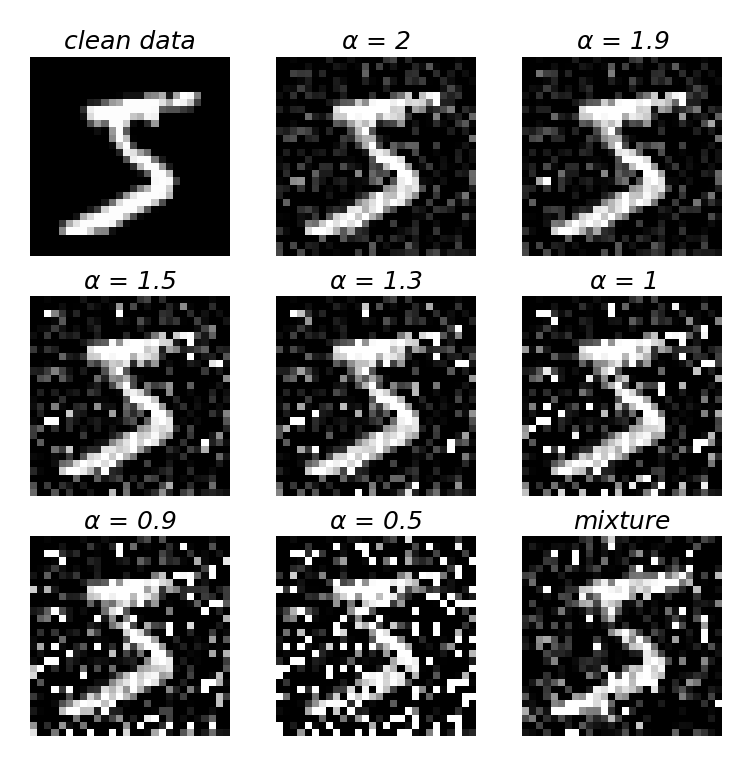}
\end{center}
\caption{An example of MNIST dataset with noise of different $\alpha$ with $\gamma=0.283$}
\label{mnist_ex}
\end{figure}

\begin{figure*}[htbp]
\begin{center}
\includegraphics{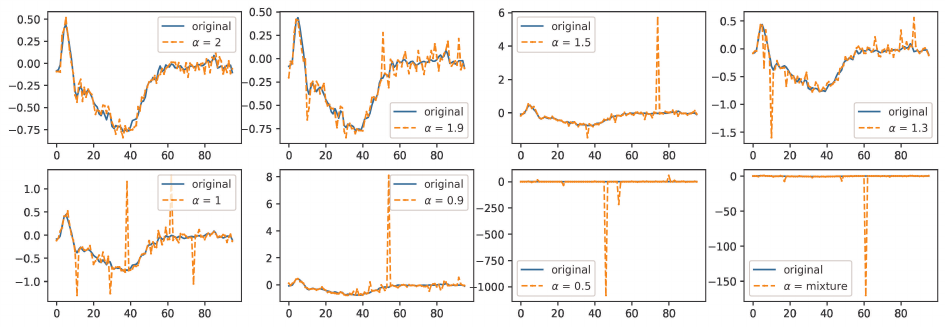}
\end{center}
\caption{An example of ECG200 dataset with noise of different $\alpha$ with $\gamma=0.021$, where the horizontal coordinates indicate the time and the vertical coordinates indicate the ECG voltage}
\label{ECG}
\end{figure*}

To create the mixture noise dataset, we replicate the dataset ten times. We then sample noise from an $\alpha$-stable mixture model \citep{salas2009finite}. The $\alpha$-stable mixture model is $f(x)=\sum_{i=1}^6 (1/6) f_i(x)$, where $f_i(x)$ denotes the $i$-th $\alpha$-stable distribution component. The six components represent $\alpha$-stable distributions with $\alpha=2,\ 1.9,\ 1.5,\ 1.3,\ 1$ and $0.9$, respectively. 
Figures \ref{mnist_ex} and \ref{ECG} depict examples illustrating the impact of incorporating $\alpha$-stable noise into the datasets. We choose only one image and one time series dataset respectively, MNIST and ECG200.

\subsection{Experiment Configuration}

To handle different types of data, we employ different models for different datasets. All the models are trained on a single RTX 3090 GPU with Python 3.6.8 and Tensorflow 2.4.1.

In this experiment, different $\alpha$-stable noises are added to four datasets (MNIST, CIFAR10, ECG200, and LIBRAS). For the MNIST dataset, we utilize an FCN  with three units in each layer and a total of three layers. In the case of classifying CIFAR10, we employ a ResNet20. For the ECG200 dataset, we utilize the VGG13 model. Lastly, for the LIBRAS dataset, we utilize LSTM model with 128 units in a single LSTM layer. We also vary the width and depth of each model, as is shown in Table \ref{parameters}, but the conclusions remain consistent. Due to space constraints, the results of the other architectures are provided in the Appendix. In order to mitigate the impact of random variations in individual results, we conduct each set of experiments in a repeated manner, repeating them five times.

\begin{table}[t]
\caption{Structure of Different Models}
\begin{center}
\begin{tabular}{ccc}		
\toprule
Model& Structure (Width, Depth)\\
\midrule
FCN & (3,3), (3,4), (3,5), (3,3), (10,3), (100,3)\\
ResNet & (8,38), (16,38), (32,38), (16,32), (16,38), (16,44)\\
VGG & (16,13), (32,13), (64,13), (128,13), (64,7), (64,9), (64,11), (64,13)\\
LSTM & (128,1), (256,1), (512,1), (128,1), (128,2), (128,3)\\
\bottomrule
\end{tabular}
\end{center}
\label{parameters}
\end{table}

\section{Results and Discussion}

\begin{figure}[htbp]
\begin{center}
\includegraphics[width=0.9\linewidth]{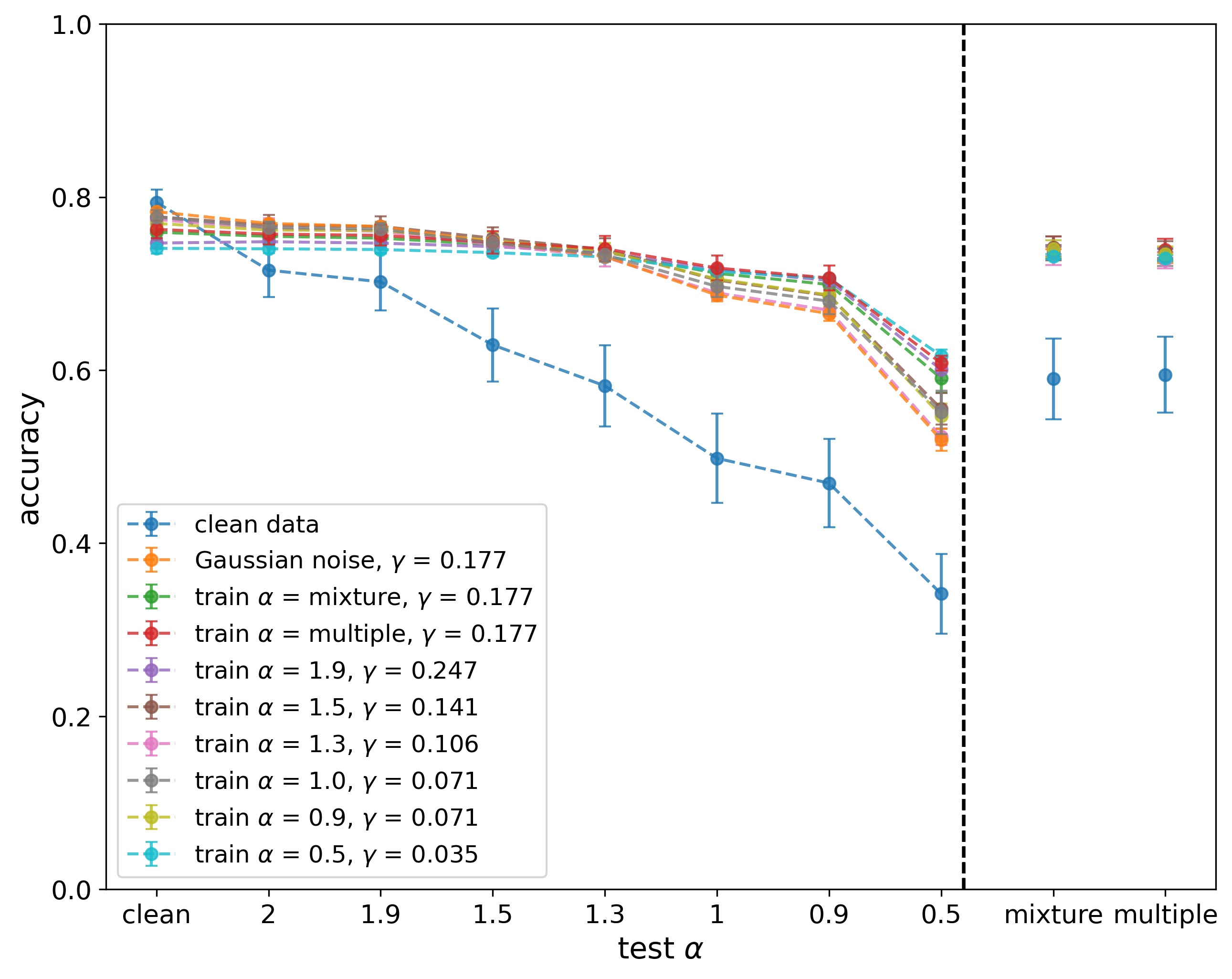} 
\end{center}
\caption{Results of FCN on MNIST of single and combined training $\alpha$}
\label{mnist2}
\end{figure}

\begin{figure}[t]
\begin{center}
\includegraphics[width=0.9\linewidth]{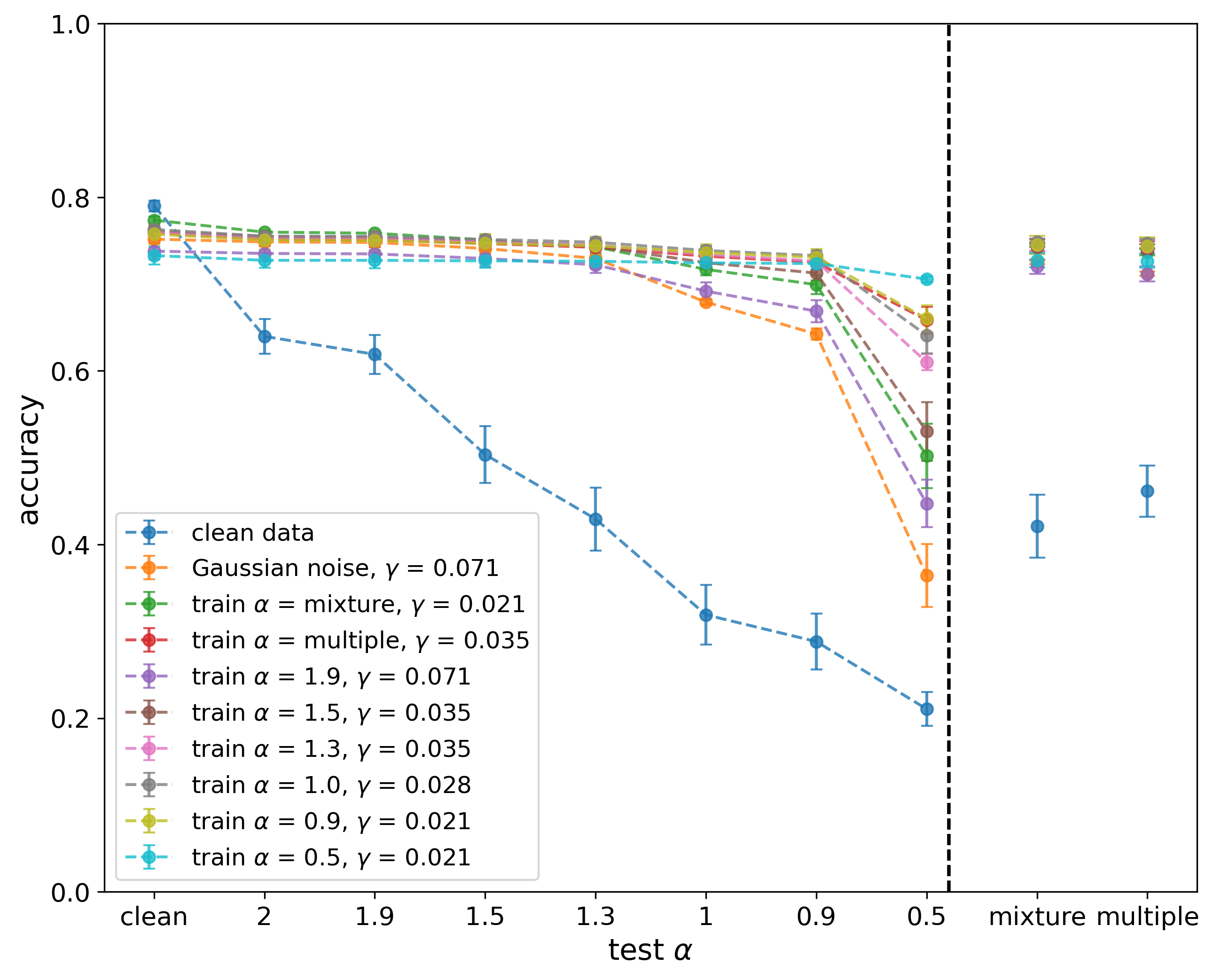}
\end{center}
\caption{Results of ResNet on CIFAR10 of single and combined training $\alpha$}
\label{cifar2}
\end{figure}

\begin{figure}[t]
\begin{center}
\includegraphics[width=0.9\linewidth]{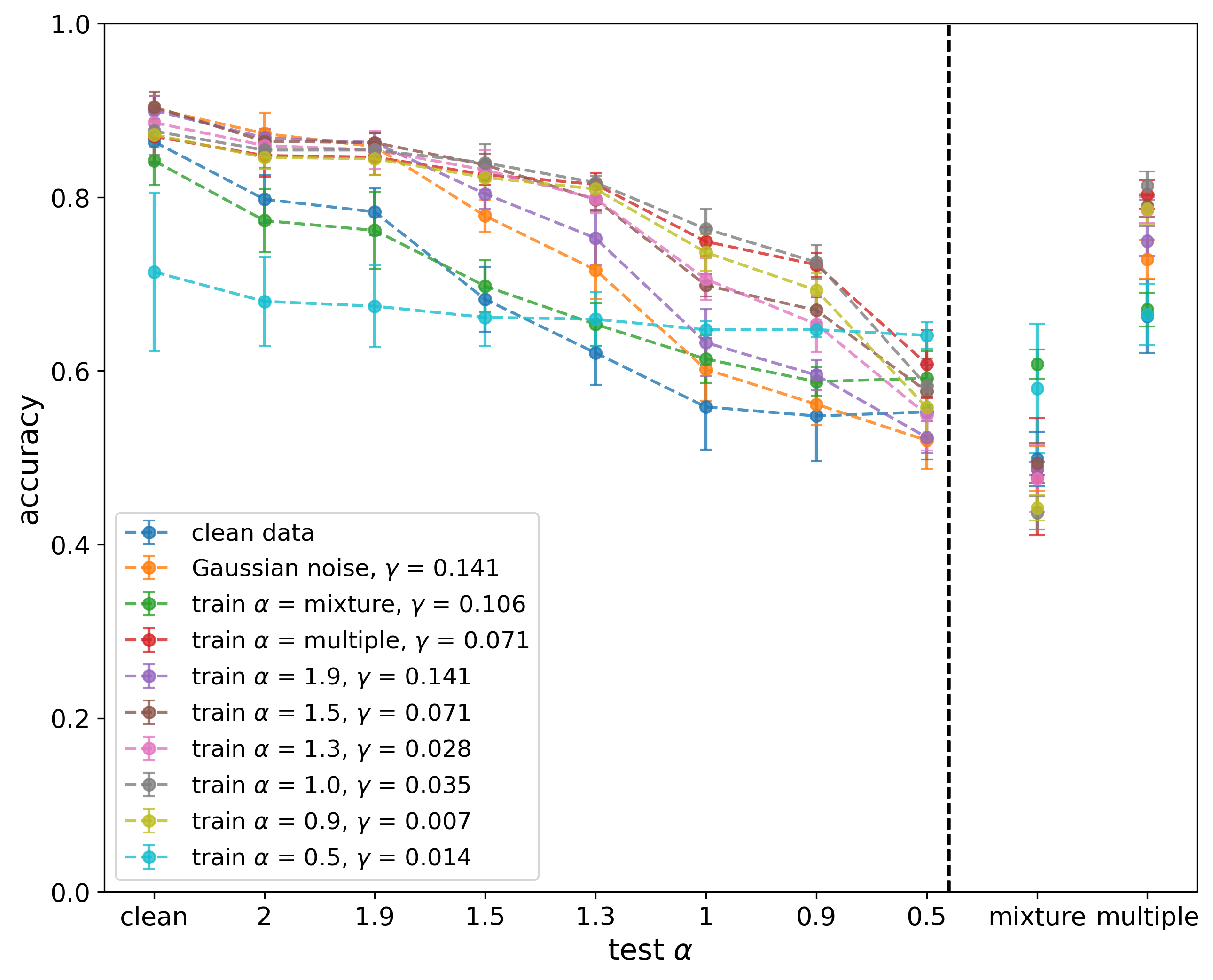} 
\end{center}
\caption{Results of VGG on ECG200 of single and combined training $\alpha$}
\label{ecg2}
\end{figure}

\begin{figure}[t]
\begin{center}
\includegraphics[width=0.9\linewidth]{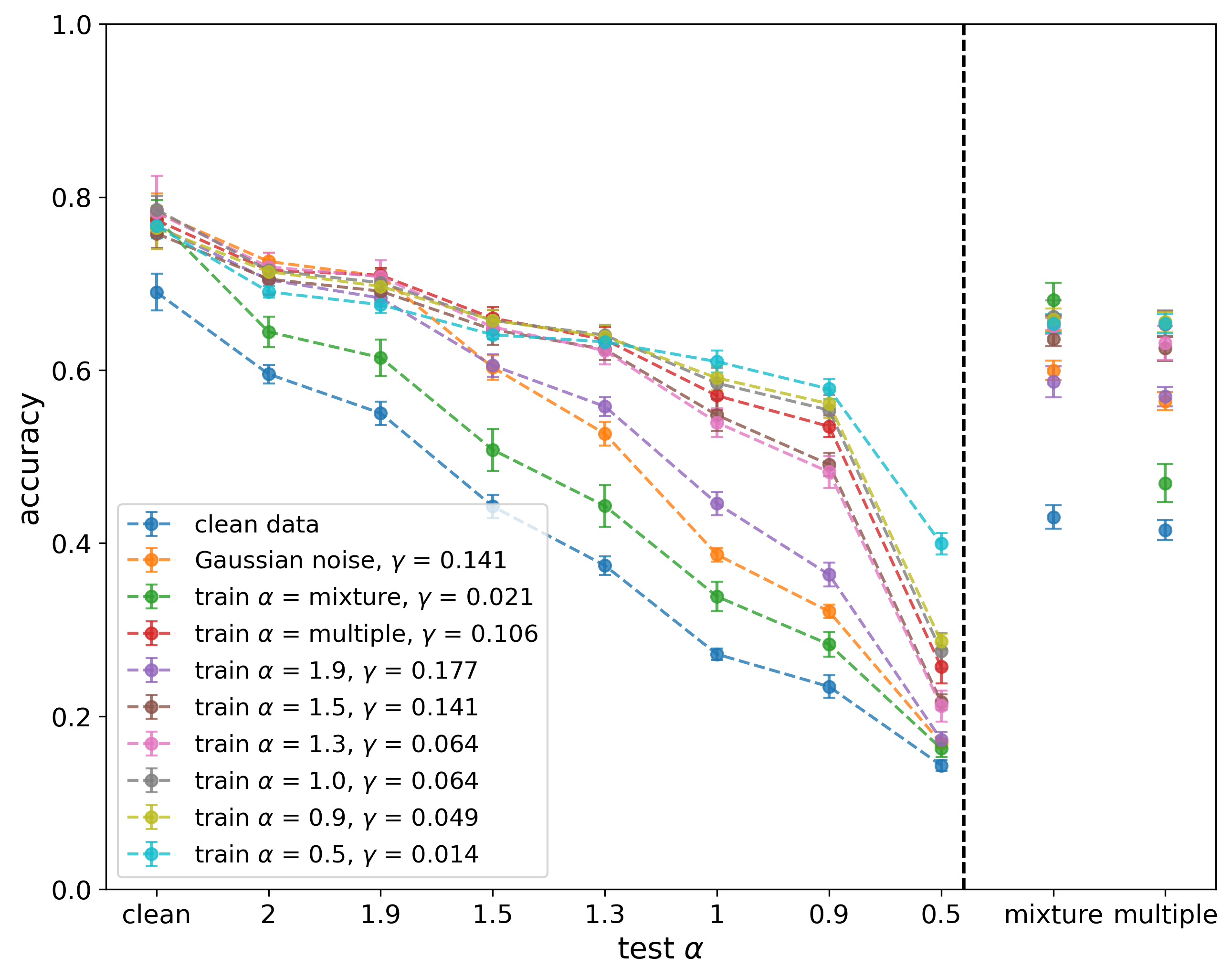} 
\end{center}
\caption{Results of LSTM on LIBRAS of single and combined training $\alpha$}
\label{libras2}
\end{figure}

\subsection{Comparison Between Different $\alpha$}

Figures \ref{mnist2} to \ref{libras2} show the performance of different $\alpha$ values in the optimal scenario, with the corresponding optimal $\gamma$ value indicated in the legend. The dots in the figure represent the mean values obtained from the five experiments, and the error bars indicate the standard deviation.

For single training $\alpha$-stable noise scenario, several common features can be observed. It is evident that the model trained solely on clean data lacks robustness in the presence of impulsive noise. Training with Gaussian noise helps alleviate this issue, although to a lesser extent compared to $\alpha$-stable noise. Notably, impulsive noise, particularly with $\alpha$ values of 0.9 and 1, substantially enhances accuracy and demonstrates relatively good performance across different testing $\alpha$ values. When the test set consists of clean data or Gaussian noise, the model trained with $\alpha$-stable noise still outperforms the others or is at least not worse than them, demonstrating its robustness on various types of data sets. This indicates that we can use $\alpha$-stable noise instead of Gaussian noise for training when we have no prior knowledge about noise. The limited prominence of training with $\alpha$-stable noise in image data compared to time series data may be due to the fact that the process of clipping the image data is analogous to truncating the noise distribution. However, even with truncation present, training with truncated $\alpha$-stable noise still yields superior results compared to truncated Gaussian noise. Truncated $\alpha$-stable and truncated Gaussian noise exhibit distinct characteristics due to their disparate algebraic and exponential decay properties, respectively. Consequently, their probability density functions (PDFs) differ significantly, even within the truncated context.

Note that the model trained with noise of $\alpha=0.5$ demonstrates excellent performance when the noise in the test set is highly impulsive, and its performance on the three datasets (MNIST, CIFAR10 and LIBRAS) when the test set noise is not impulsive is still reasonably good, albeit with a slight decrease in accuracy. It is worth mentioning that datasets with highly impulsive noise are inherently extreme and rare in nature, and the model's performance on the ECG200 dataset is not stable when training with noise of $\alpha=0.5$. Therefore, choosing a training noise level of $\alpha=0.5$ is risky, and we do not recommend doing so. Instead, we recommend selecting a noise level of approximately $\alpha=1$ for data augmentation during training, as it exhibits high accuracy on the majority of the test set.
\par
In the case of models trained with combined-noise, it is observed that multiple noise training outperforms mixture noise training. This phenomenon can be attributed to the fact that data trained with multiple noise types can more effectively capture the underlying data structure when each type of noise is introduced individually.
\par
 Our experimental results indicate that, for the case of single-noise, the optimal $\alpha$ value for most datasets is 1. For the case of combined-noise, the commonly observed optimal solution is the multiple noise. In Table \ref{improve1} and \ref{improve2}, we present the percentage improvement on different test datasets. This improvement is calculated after training the model with noise $\alpha=1$ and multiple noise, compared to the model trained with clean data and Gaussian noise. It is evident that training the model with $\alpha$-stable noise leads to an improvement in its average accuracy across a broad spectrum of noise scenarios. Furthermore, this approach appears to perform better on complex datasets. We propose that data augmentation with Cauchy noise and multiple $\alpha$-stable noise are more effective in enhancing model robustness compared to Gaussian noise. Furthermore, since we have conducted experiments on a diverse range of datasets including images and time series, our approach can be considered generalizable.

\begin{table}[h]
\caption{Percentage Improvement of Model Trained with Noise of $\alpha = 1$}
\label{improve1}
\begin{center}
\begin{tabular}{cccccc}
   \toprule
   Baseline & FCN & ResNet & VGG & LSTM\\
   \midrule
    Clean & 21.36\% & 57.63\% & 15.14\% & 50.24\% \\
    Gaussian & 0.65\% & 7.88\% & 7.62\% & 15.67\% \\
   \bottomrule
\end{tabular}
\end{center}
\end{table}

\begin{table}[h]
\caption{Percentage Improvement of Model Trained with Multiple Noise}
\label{improve2}
\begin{center}
\begin{tabular}{cccccc}
   \toprule
   Baseline & FCN & ResNet & VGG & LSTM\\
   \midrule
    Clean & 22.99\% & 57.01\% & 15.18\% & 48.64\% \\
    Gaussian & 2.01\% & 7.45\% & 7.65\% & 14.43\% \\
   \bottomrule
\end{tabular}
\end{center}
\end{table}

\subsection{Results on Benchmark Datasets}

We conduct experiments on two widely recognized benchmark corrupted datasets, namely MNIST-C and CIFAR10-C. These datasets consist of corrupted versions of MNIST and CIFAR10, respectively, and encompass a diverse range of prevalent perturbations, such as Gaussian noise, impulse noise, blur, and various weather conditions. Each corruption type has a corresponding subdataset. The primary objective of this investigation is to explore the applicability of the $\alpha$-stable data augmentation method to different types of corruptions, extending beyond impulsive noise. We compute the test accuracies of models trained with Gaussian noise, Cauchy noise, and multiple noise on all subdatasets, alongside the average accuracy across all subdatasets. For the sake of brevity, we present a subset of the corruption types in Table \ref{mnist_pretrain} and \ref{cifar_pretrain}. As evident from previous studies \citep{lecun1989optimal,hassibi1992second,molchanov2017variational,hoefler2021sparsity}, incorporating sparsity in a model has been observed to enhance its generalization capability. Consequently, we compute the sparsity of models trained using various data augmentation techniques to facilitate a comparative analysis. By examining the parameter distributions through their associated histograms, we observe a substantial concentration of model parameters around the value of 0. Accordingly, we establish a threshold of 0.01 and calculate the proportion of parameters with absolute values falling below this threshold. The results are shown in the last column of Table \ref{mnist_pretrain} and \ref{cifar_pretrain}.

\begin{table*}[htp]
\caption{Average Accuracy Across the Entire MNIST-C Dataset, Accuracy of Various Models on Different Representative Corruption Types and Model Sparsity}
\label{mnist_pretrain}
\begin{center}
\begin{tabular}{c|cccccc|c}
   \bottomrule
       &  \multicolumn{6}{c|}{Accuracy} & \\
   Model    &Average & Impulse Noise & Spatter & Scale & Dotted Line & Zig Zag & Model Sparsity\\
   \hline
    Clean & 42.61\%  & 39.31\%  & 58.53\%& 27.21\% & 64.54\% & 34.87\% & 3.87\%\\
    Gaussian noise  & 45.59\%  & 57.46\%& 67.57\% & 30.48\%  & 66.64\% & 47.00\% & 3.02\%\\
    Multiple noise & \textbf{51.46\%}  & \textbf{72.01\%} & \textbf{73.66\%} & \textbf{35.98\%} & \textbf{71.52\%} & \textbf{59.72\%} & 6.16\% \\
    Cauchy noise  & 46.32\%  & 66.61\% & 71.80\% & 23.51\% & 70.62\% & 57.97\% & \textbf{7.13\%}\\
   \toprule
\end{tabular}
\end{center}
\end{table*}

\begin{table*}[htp]
\caption{Average Accuracy Across the Entire CIFAR10-C Dataset, Accuracy of Various Models on Different Representative Corruption Types and Model Sparsity}
\label{cifar_pretrain}
\begin{center}
\begin{tabular}{c|cccccc|c}
   \bottomrule
       &  \multicolumn{6}{c|}{Accuracy} & \\
   Model  & Average & \makecell{Impulse noise} & Spatter & \makecell{Gaussian\\Blur} & \makecell{Defocus\\Blur} & \makecell{Glass\\Blur} & \makecell{Model Sparsity}\\
   \hline
    Clean & 60.45\%  & 49.11\% & 67.40\%  & 60.68\%& 66.86\% & 33.44\% & 47.05\% \\
    Gaussian noise  & 65.29\%  & 62.99\%& 67.38\% & 60.91\%  & 64.69\% &  58.31\% & 51.62\%\\
    Multiple noise & 68.46\%  & 74.23\% & 70.99\% & \textbf{67.85\%} & \textbf{70.25\%} & 65.45\% & 54.60\% \\
    Cauchy noise  & \textbf{68.91\%}  & \textbf{75.56\%} & \textbf{71.57\%} & 66.86\% & 69.40\%  & \textbf{66.63\%} & \textbf{59.49\%} \\
   \toprule
\end{tabular}
\end{center}
\end{table*}

As observed in the results, the utilization of $\alpha$-stable noise for data augmentation not only enhances the model's resilience to impulsive noise within the data but also surpasses the performance of the model trained with Gaussian noise when confronted with other corruption types. This outcome substantiates the general applicability and effectiveness of our approach. Models trained with $\alpha$-stable noise exhibit a higher degree of sparsity compared to those trained with Gaussian noise, indicating that our model likely eliminates redundant features. This further confirms the stronger generalization capability of our model. Due to limited datasets availability, our experiments have solely focused on alternative corruption types within image datasets. As a result, the generalization performance of $\alpha$-stable noise across other data types warrants further investigation and discussion.

\section{CONCLUSION AND FUTURE WORK} 
\label{sec:conclusion}
From our experimental findings, it can be concluded that the performance of data augmentation utilizing Cauchy noise and multiple $\alpha$-stable noise outperform Gaussian noise across a wide range of noise scenarios, particularly when the dataset is corrupted by impulsive noise. Moreover, it is not inferior to Gaussian trained models even for clean data or data with Gaussian noise. Therefore, in situations where a prior knowledge of the noise in dataset is unavailable, employing $\alpha$-stable noise is, at the very least, as effective as using Gaussian noise. Given our extensive testing across various contexts, including images, time series, and corrupted datasets, our results exhibit a high degree of generalizability. Our findings indicate that the specific model structure has minimal impact on the influence of noise, further underscoring the stability of our approach as well.
\par
The variations in dataset sizes and distributions lead to different levels of suitable noise intensities. While the use of multiple noise and setting $\alpha=1$ has shown improved results in different datasets, determining the optimal $\gamma$ value for different datasets poses a challenging task that requires attention. In a recent study \citep{ning2021improving}, an adaptive approach is proposed to dynamically adjust noise perturbation levels for individual examples during training. Considering that modifying the training $\gamma$ can be seen as altering the noise perturbation level, a similar approach could be considered in our future research.

Our study explores the effectiveness of $\alpha$-stable noise data augmentation in general classification tasks and finds that it demonstrates greater superiority when the data is corrupted by impulsive noise. Therefore, the application of this method in datasets where impulsive noise is frequently present, such as radar and sonar data, as well as medical data like ECG and MRI, holds promising prospects. Furthermore, while our current research primarily centers around image and time series data, it is imperative to acknowledge that noise exerts a substantial influence on audio, video, text, and various other forms of data in everyday life. Consequently, this work lends itself to future extensions encompassing diverse data types, facilitating an exploration of the impact of $\alpha$-stable noise on each domain. Additionally, we have only explored the role of $\alpha$-stable noise in classification tasks so far. In the future, it is worth investigating its impact on other supervised learning tasks such as regression, object detection, and more.

\bibliography{ourpaper1}

\begin{thebibliography}{}

\bibitem[Adilova et~al., 2019]{DBLP:conf/pkdd/AdilovaPS18}
Adilova, L., Paul, N., and Schlicht, P. (2019).
\newblock Introducing noise in decentralized training of neural networks.
\newblock In {\em ECML PKDD 2018 Workshops: DMLE 2018 and IoTStream 2018, Dublin, Ireland, September 10-14, 2018, Revised Selected Papers 18}, pages 37--48. Springer.

\bibitem[Ahn, 2020]{ahn2020neural}
Ahn, K.-H. (2020).
\newblock A neural network ensemble approach with jittered basin characteristics for regionalized low flow frequency analysis.
\newblock {\em Journal of Hydrology}, 590:125501.

\bibitem[An, 1996]{DBLP:journals/neco/An96}
An, G. (1996).
\newblock The effects of adding noise during backpropagation training on a generalization performance.
\newblock {\em Neural computation}, 8(3):643--674.

\bibitem[Aubry et~al., 2016]{7472965}
Aubry, A., Maio, A.~D., Carotenuto, V., and Farina, A. (2016).
\newblock Radar phase noise modeling and effects-part i : Mti filters.
\newblock {\em IEEE Transactions on Aerospace and Electronic Systems}, 52(2):698--711.

\bibitem[Audhkhasi et~al., 2013]{DBLP:conf/ijcnn/AudhkhasiOK13}
Audhkhasi, K., Osoba, O., and Kosko, B. (2013).
\newblock Noise benefits in backpropagation and deep bidirectional pre-training.
\newblock In {\em The 2013 International Joint Conference on Neural Networks (IJCNN)}, pages 1--8. IEEE.

\bibitem[Berger and Mandelbrot, 1963]{Mandelbrot1963}
Berger, J.~M. and Mandelbrot, B. (1963).
\newblock A new model for error clustering in telephone circuits.
\newblock {\em IBM Journal of Research and Development}, 7(3):224--236.

\bibitem[Bishop, 1995]{bishop1995training}
Bishop, C.~M. (1995).
\newblock Training with noise is equivalent to tikhonov regularization.
\newblock {\em Neural computation}, 7(1):108--116.

\bibitem[Dua and Graff, 2017]{Dua:2019}
Dua, D. and Graff, C. (2017).
\newblock {UCI} machine learning repository.

\bibitem[Fawzi et~al., 2018]{DBLP:journals/ml/FawziFF18}
Fawzi, A., Fawzi, O., and Frossard, P. (2018).
\newblock Analysis of classifiers’ robustness to adversarial perturbations.
\newblock {\em Machine learning}, 107(3):481--508.

\bibitem[Fawzi et~al., 2016]{DBLP:conf/nips/FawziMF16}
Fawzi, A., Moosavi-Dezfooli, S.-M., and Frossard, P. (2016).
\newblock Robustness of classifiers: from adversarial to random noise.
\newblock {\em Advances in neural information processing systems}, 29.

\bibitem[Grandvalet et~al., 1997]{DBLP:journals/neco/GrandvaletCB97}
Grandvalet, Y., Canu, S., and Boucheron, S. (1997).
\newblock Noise injection: Theoretical prospects.
\newblock {\em Neural Computation}, 9(5):1093--1108.

\bibitem[Hassibi and Stork, 1992]{hassibi1992second}
Hassibi, B. and Stork, D. (1992).
\newblock Second order derivatives for network pruning: Optimal brain surgeon.
\newblock {\em Advances in neural information processing systems}, 5.

\bibitem[Hendrycks and Dietterich, 2019]{hendrycks2019benchmarking}
Hendrycks, D. and Dietterich, T. (2019).
\newblock Benchmarking neural network robustness to common corruptions and perturbations.
\newblock {\em arXiv preprint arXiv:1903.12261}.

\bibitem[Hoefler et~al., 2021]{hoefler2021sparsity}
Hoefler, T., Alistarh, D., Ben-Nun, T., Dryden, N., and Peste, A. (2021).
\newblock Sparsity in deep learning: Pruning and growth for efficient inference and training in neural networks.
\newblock {\em The Journal of Machine Learning Research}, 22(1):10882--11005.

\bibitem[Holmstrom et~al., 1992]{holmstrom1992using}
Holmstrom, L., Koistinen, P., et~al. (1992).
\newblock Using additive noise in back-propagation training.
\newblock {\em IEEE transactions on neural networks}, 3(1):24--38.

\bibitem[Huang, 2008]{4738222}
Huang, T. (2008).
\newblock Prior training with jittered series for time series forecasting.
\newblock In {\em 2008 IEEE International Conference on Industrial Engineering and Engineering Management}, pages 2001--2005.

\bibitem[Isaev et~al., 2018]{DBLP:conf/icann/IsaevBDLVD18}
Isaev, I., Burikov, S., Dolenko, T., Laptinskiy, K., Vervald, A., and Dolenko, S. (2018).
\newblock Joint application of group determination of parameters and of training with noise addition to improve the resilience of the neural network solution of the inverse problem in spectroscopy to noise in data.
\newblock In {\em Artificial Neural Networks and Machine Learning--ICANN 2018: 27th International Conference on Artificial Neural Networks, Rhodes, Greece, October 4-7, 2018, Proceedings, Part I 27}, pages 435--444. Springer.

\bibitem[Isaev and Dolenko, 2016]{isaev2016training}
Isaev, I. and Dolenko, S. (2016).
\newblock Training with noise as a method to increase noise resilience of neural network solution of inverse problems.
\newblock {\em Optical Memory and Neural Networks}, 25:142--148.

\bibitem[Kalavathi and Priya, 2016]{Kalavathi_Priya_2016}
Kalavathi, P. and Priya, T. (2016).
\newblock Removal of impulse noise using histogram-based localized wiener filter for mr brain image restoration.
\newblock In {\em 2016 IEEE International Conference on Advances in Computer Applications (ICACA)}, pages 4--8.

\bibitem[Karaku{\c{s}} et~al., 2018]{karakucs2018generalized}
Karaku{\c{s}}, O., Kuruo{\u{g}}lu, E.~E., and Alt{\i}nkaya, M.~A. (2018).
\newblock Generalized bayesian model selection for speckle on remote sensing images.
\newblock {\em IEEE Transactions on Image Processing}, 28(4):1748--1758.

\bibitem[Karaku{\c{s}} et~al., 2020]{DBLP:journals/sivp/KarakusKA20}
Karaku{\c{s}}, O., Kuruo{\u{g}}lu, E.~E., and Alt{\i}nkaya, M.~A. (2020).
\newblock Modelling impulsive noise in indoor powerline communication systems.
\newblock {\em Signal, image and video processing}, 14(8):1655--1661.

\bibitem[Kosko et~al., 2020]{DBLP:journals/nn/KoskoAO20}
Kosko, B., Audhkhasi, K., and Osoba, O. (2020).
\newblock Noise can speed backpropagation learning and deep bidirectional pretraining.
\newblock {\em Neural Networks}, 129:359--384.

\bibitem[Krizhevsky, 2009]{Krizhevsky09learningmultiple}
Krizhevsky, A. (2009).
\newblock Learning multiple layers of features from tiny images.
\newblock Technical report.

\bibitem[Larochelle et~al., 2007]{DBLP:conf/icml/LarochelleECBB07}
Larochelle, H., Erhan, D., Courville, A., Bergstra, J., and Bengio, Y. (2007).
\newblock An empirical evaluation of deep architectures on problems with many factors of variation.
\newblock In {\em Proceedings of the 24th international conference on Machine learning}, pages 473--480.

\bibitem[LeCun and Cortes, 2010]{lecun-mnisthandwrittendigit-2010}
LeCun, Y. and Cortes, C. (2010).
\newblock {MNIST} handwritten digit database.

\bibitem[LeCun et~al., 1989]{lecun1989optimal}
LeCun, Y., Denker, J., and Solla, S. (1989).
\newblock Optimal brain damage.
\newblock {\em Advances in neural information processing systems}, 2.

\bibitem[Lee et~al., 2023]{lee2023deep}
Lee, W., Nam, H.~S., Seok, J.~Y., Oh, W.-Y., Kim, J.~W., and Yoo, H. (2023).
\newblock Deep learning-based image enhancement in optical coherence tomography by exploiting interference fringe.
\newblock {\em Communications Biology}, 6(1):464.

\bibitem[Lévy, 1937]{levy1937theorie}
Lévy, P. (1937).
\newblock {\em Theorie de l'addition des variables aleatoires [Combination theory of unpredictable variables]}.
\newblock Gauthier-Villars, Paris.

\bibitem[Matsuoka, 1992]{DBLP:journals/tsmc/Matsuoka92}
Matsuoka, K. (1992).
\newblock Noise injection into inputs in back-propagation learning.
\newblock {\em IEEE Transactions on Systems, Man, and Cybernetics}, 22(3):436--440.

\bibitem[Molchanov et~al., 2017]{molchanov2017variational}
Molchanov, D., Ashukha, A., and Vetrov, D. (2017).
\newblock Variational dropout sparsifies deep neural networks.
\newblock In {\em International Conference on Machine Learning}, pages 2498--2507. PMLR.

\bibitem[Moosavi-Dezfooli et~al., 2017]{DBLP:conf/cvpr/Moosavi-Dezfooli17}
Moosavi-Dezfooli, S.-M., Fawzi, A., Fawzi, O., and Frossard, P. (2017).
\newblock Universal adversarial perturbations.
\newblock In {\em Proceedings of the IEEE conference on computer vision and pattern recognition}, pages 1765--1773.

\bibitem[Moreno-Barea et~al., 2018]{DBLP:conf/ssci/Moreno-BareaSJU18}
Moreno-Barea, F.~J., Strazzera, F., Jerez, J.~M., Urda, D., and Franco, L. (2018).
\newblock Forward noise adjustment scheme for data augmentation.
\newblock In {\em 2018 IEEE symposium series on computational intelligence (SSCI)}, pages 728--734. IEEE.

\bibitem[Mu and Gilmer, 2019]{mu2019mnist}
Mu, N. and Gilmer, J. (2019).
\newblock Mnist-c: A robustness benchmark for computer vision.
\newblock {\em arXiv preprint arXiv:1906.02337}.

\bibitem[Ning et~al., 2021]{ning2021improving}
Ning, K.-P., Tao, L., Chen, S., and Huang, S.-J. (2021).
\newblock Improving model robustness by adaptively correcting perturbation levels with active queries.
\newblock In {\em Proceedings of the AAAI Conference on Artificial Intelligence}, volume~35, pages 9161--9169.

\bibitem[Ochoa{-}Brust et~al., 2019]{DBLP:journals/informaticaSI/Ochoa-BrustMFGM19}
Ochoa{-}Brust, A.~M., Mena, L.~J., Felix, V.~G., Gonz{\'{a}}lez, A., Mata{-}L{\'{o}}pez, W.~A., and Maestre, G. (2019).
\newblock Noise-tolerant modular neural network system for classifying {ECG} signal.
\newblock {\em Informatica (Slovenia)}, 43(1).

\bibitem[Olszewski et~al., 2001]{10.5555/935627}
Olszewski, R.~T., Maxion, R., and Siewiorek, D. (2001).
\newblock {\em Generalized Feature Extraction for Structural Pattern Recognition in Time-Series Data}.
\newblock PhD thesis, USA.
\newblock AAI3040489.

\bibitem[Plaut et~al., 1986]{plaut1986experiments}
Plaut, D.~C. et~al. (1986).
\newblock Experiments on learning by back propagation.

\bibitem[Reed et~al., 1992]{reed1992regularization}
Reed, R., Oh, S., Marks, R., et~al. (1992).
\newblock Regularization using jittered training data.
\newblock In {\em International joint conference on neural networks}, volume~3, pages 147--152.

\bibitem[Reyes and Duro, 2001]{DBLP:journals/ijon/ReyesD01}
Reyes, J.~S. and Duro, R.~J. (2001).
\newblock Influence of noise on discrete time backpropagation trained networks.
\newblock {\em Neurocomputing}, 41(1-4):67--89.

\bibitem[Rusak et~al., 2020a]{DBLP:journals/corr/abs-2001-06057}
Rusak, E., Schott, L., Zimmermann, R.~S., Bitterwolf, J., Bringmann, O., Bethge, M., and Brendel, W. (2020a).
\newblock Increasing the robustness of dnns against image corruptions by playing the game of noise.
\newblock {\em CoRR}, abs/2001.06057.

\bibitem[Rusak et~al., 2020b]{Rusak_2020}
Rusak, E., Schott, L., Zimmermann, R.~S., Bitterwolf, J., Bringmann, O., Bethge, M., and Brendel, W. (2020b).
\newblock {\em A simple way to make neural networks robust against diverse image corruptions}, page 53–69.

\bibitem[Salas-Gonzalez et~al., 2009]{salas2009finite}
Salas-Gonzalez, D., Kuruoglu, E.~E., and Ruiz, D.~P. (2009).
\newblock Finite mixture of $\alpha$-stable distributions.
\newblock {\em Digital Signal Processing}, 19(2):250--264.

\bibitem[Samorodnitsky and Taqqu, 1994]{samorodnitsky1994stable}
Samorodnitsky, G. and Taqqu, M.~S. (1994).
\newblock {\em Stable non-gaussian random processes: stochastic models with infinite variance}.
\newblock New York: Chapman-Hall.

\bibitem[Seghouane et~al., 2002]{DBLP:conf/nnsp/SeghouaneMF02}
Seghouane, A.-K., Moudden, Y., and Fleury, G. (2002).
\newblock On learning feedforward neural networks with noise injection into inputs.
\newblock In {\em Proceedings of the 12th IEEE Workshop on Neural Networks for Signal Processing}, pages 149--158. IEEE.

\bibitem[Shao and Nikias, 1993]{DBLP:journals/pieee/ShaoN93}
Shao, M. and Nikias, C.~L. (1993).
\newblock Signal processing with fractional lower order moments: stable processes and their applications.
\newblock {\em Proceedings of the IEEE}, 81(7):986--1010.

\bibitem[Shen et~al., 2015]{shen2015observation}
Shen, X., Zhang, H., Xu, Y., and Meng, S. (2015).
\newblock Observation of alpha-stable noise in the laser gyroscope data.
\newblock {\em IEEE Sensors Journal}, 16(7):1998--2003.

\bibitem[Sietsma and Dow, 1988]{sietsma1988neural}
Sietsma and Dow (1988).
\newblock Neural net pruning-why and how.
\newblock In {\em IEEE 1988 international conference on neural networks}, pages 325--333. IEEE.

\bibitem[Simsekli et~al., 2019]{DBLP:journals/corr/abs-1912-00018}
Simsekli, U., G{\"{u}}rb{\"{u}}zbalaban, M., Nguyen, T.~H., Richard, G., and Sagun, L. (2019).
\newblock On the heavy-tailed theory of stochastic gradient descent for deep neural networks.
\newblock {\em CoRR}, abs/1912.00018.

\bibitem[Singh et~al., 2017]{Singh_Bhole_Sharma_2017}
Singh, P., Bhole, K., and Sharma, A. (2017).
\newblock Adaptive filtration techniques for impulsive noise removal from ecg.
\newblock In {\em 2017 14th IEEE India Council International Conference (INDICON)}, pages 1--4.

\bibitem[Srinivasan et~al., 2019]{DBLP:conf/eusipco/SrinivasanKMSN19}
Srinivasan, V., Kuruoglu, E.~E., Müller, K.-R., Samek, W., and Nakajima, S. (2019).
\newblock Black-box decision based adversarial attack with symmetric {\(\alpha\)}-stable distribution.
\newblock In {\em 2019 27th European Signal Processing Conference (EUSIPCO)}, pages 1--5.

\bibitem[Stuck and Kleiner, 1974]{stuck1974statistical}
Stuck, B.~W. and Kleiner, B. (1974).
\newblock A statistical analysis of telephone noise.
\newblock {\em Bell System Technical Journal}, 53(7):1263--1320.

\bibitem[Szegedy et~al., 2013]{DBLP:journals/corr/SzegedyZSBEGF13}
Szegedy, C., Zaremba, W., Sutskever, I., Bruna, J., Erhan, D., Goodfellow, I., and Fergus, R. (2013).
\newblock Intriguing properties of neural networks.
\newblock {\em arXiv preprint arXiv:1312.6199}.

\bibitem[Venton et~al., 2021]{venton2021robustness}
Venton, J., Harris, P.~M., Sundar, A., Smith, N.~A., and Aston, P.~J. (2021).
\newblock Robustness of convolutional neural networks to physiological electrocardiogram noise.
\newblock {\em Philosophical Transactions of the Royal Society A}, 379(2212):20200262.

\bibitem[Windyga, 2001]{windyga2001fast}
Windyga, P.~S. (2001).
\newblock Fast impulsive noise removal.
\newblock {\em IEEE transactions on image processing}, 10(1):173--179.

\bibitem[Yin et~al., 2015]{DBLP:journals/ejasmp/YinLZLWTZL15}
Yin, S., Liu, C., Zhang, Z., Lin, Y., Wang, D., Tejedor, J., Zheng, T.~F., and Li, Y. (2015).
\newblock Noisy training for deep neural networks in speech recognition.
\newblock {\em EURASIP Journal on Audio, Speech, and Music Processing}, 2015:1--14.

\bibitem[Zantedeschi et~al., 2017]{zantedeschi2017efficient}
Zantedeschi, V., Nicolae, M., and Rawat, A. (2017).
\newblock Efficient defenses against adversarial attacks.
\newblock In Thuraisingham, B., Biggio, B., Freeman, D.~M., Miller, B., and Sinha, A., editors, {\em Proceedings of the 10th {ACM} Workshop on Artificial Intelligence and Security, AISec@CCS 2017, Dallas, TX, USA, November 3, 2017}, pages 39--49. {ACM}.

\bibitem[Zhan et~al., 2019]{zhan2019recovery}
Zhan, C., Yan, M., and Hao, D. (2019).
\newblock Recovery performance of lplq-admm algorithm under s$\alpha$s impulse noise.
\newblock In {\em 2019 IEEE-APS Topical Conference on Antennas and Propagation in Wireless Communications (APWC)}, pages 079--084. IEEE.

\bibitem[Zhang, 2007]{DBLP:journals/isci/Zhang07}
Zhang, G.~P. (2007).
\newblock A neural network ensemble method with jittered training data for time series forecasting.
\newblock {\em Information Sciences}, 177(23):5329--5346.

\bibitem[Zur et~al., 2009]{zur2009noise}
Zur, R.~M., Jiang, Y., Pesce, L.~L., and Drukker, K. (2009).
\newblock Noise injection for training artificial neural networks: A comparison with weight decay and early stopping.
\newblock {\em Medical physics}, 36(10):4810--4818.

\end{thebibliography}

\end{document}


%

%

\onecolumn
\aistatstitle{Supplementary Materials}

\section{Results of FCN on MNIST}

\begin{figure}[htbp]
\centering
\begin{minipage}[t]{0.48\textwidth}
\centering
\includegraphics[width=3in]{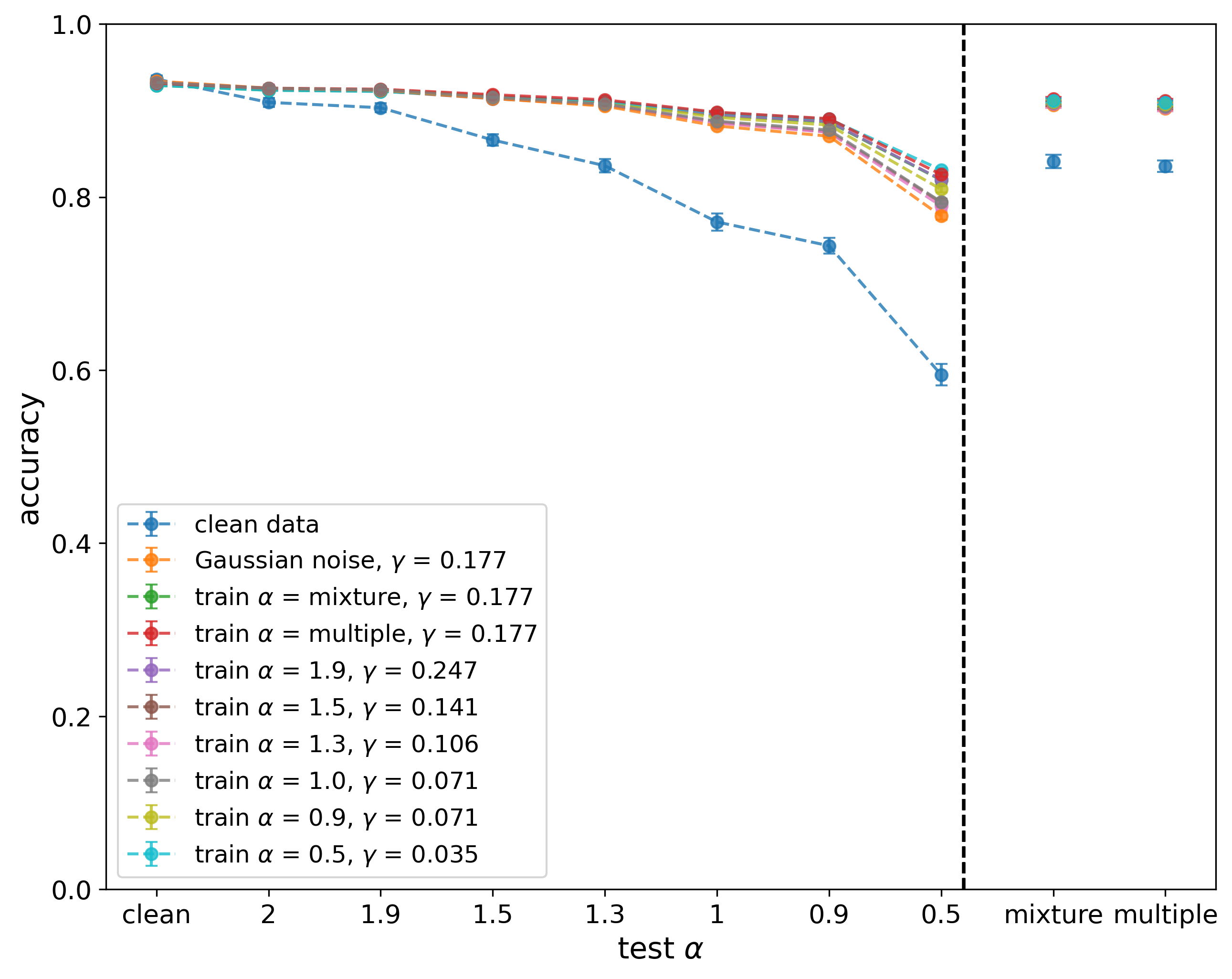}
\caption{width=10, depth=3}
\end{minipage}
\begin{minipage}[t]{0.48\textwidth}
\centering
\includegraphics[width=3in]{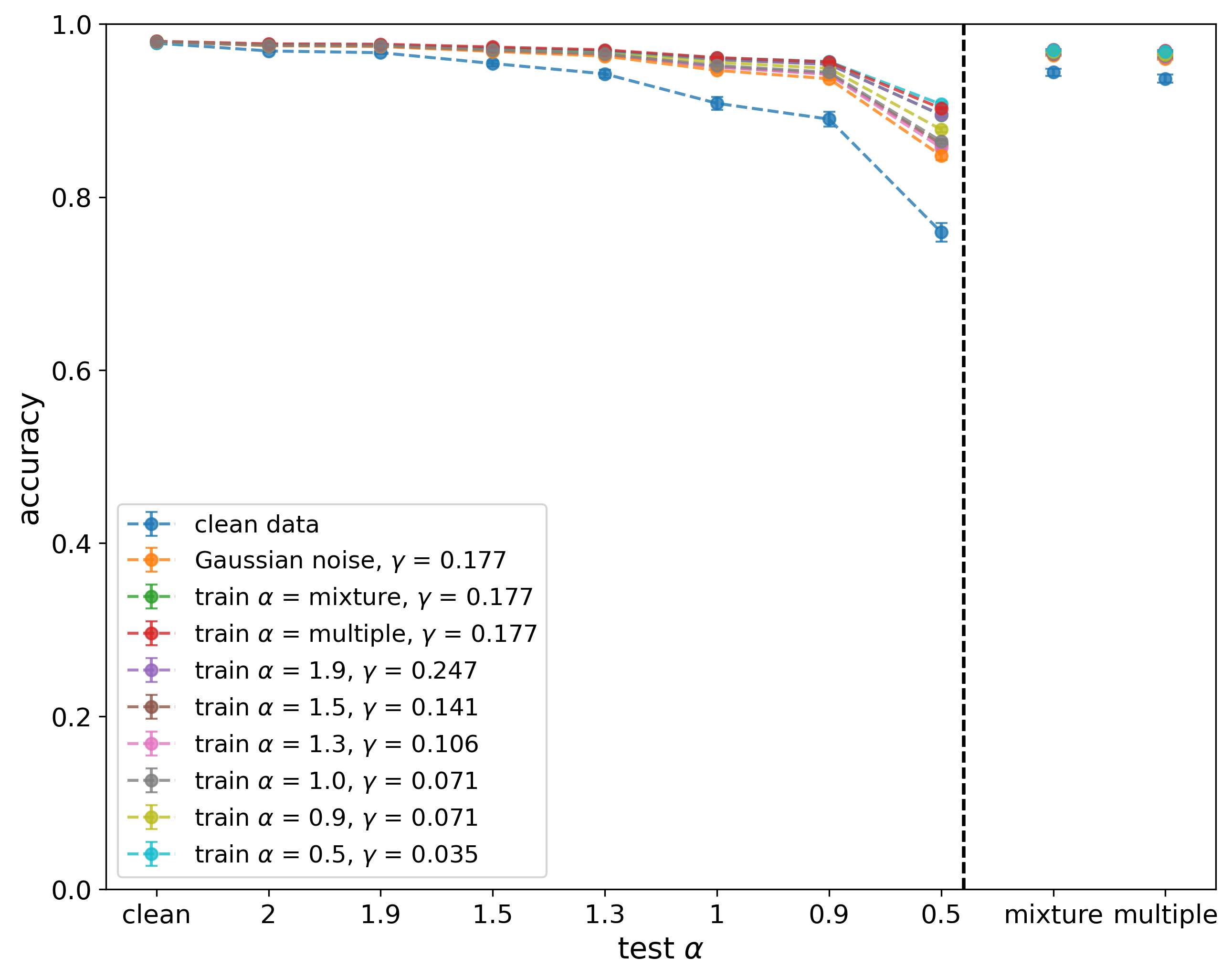}
\caption{width=100, depth=3}
\end{minipage}
\begin{minipage}[t]{0.48\textwidth}
\centering
\includegraphics[width=3in]{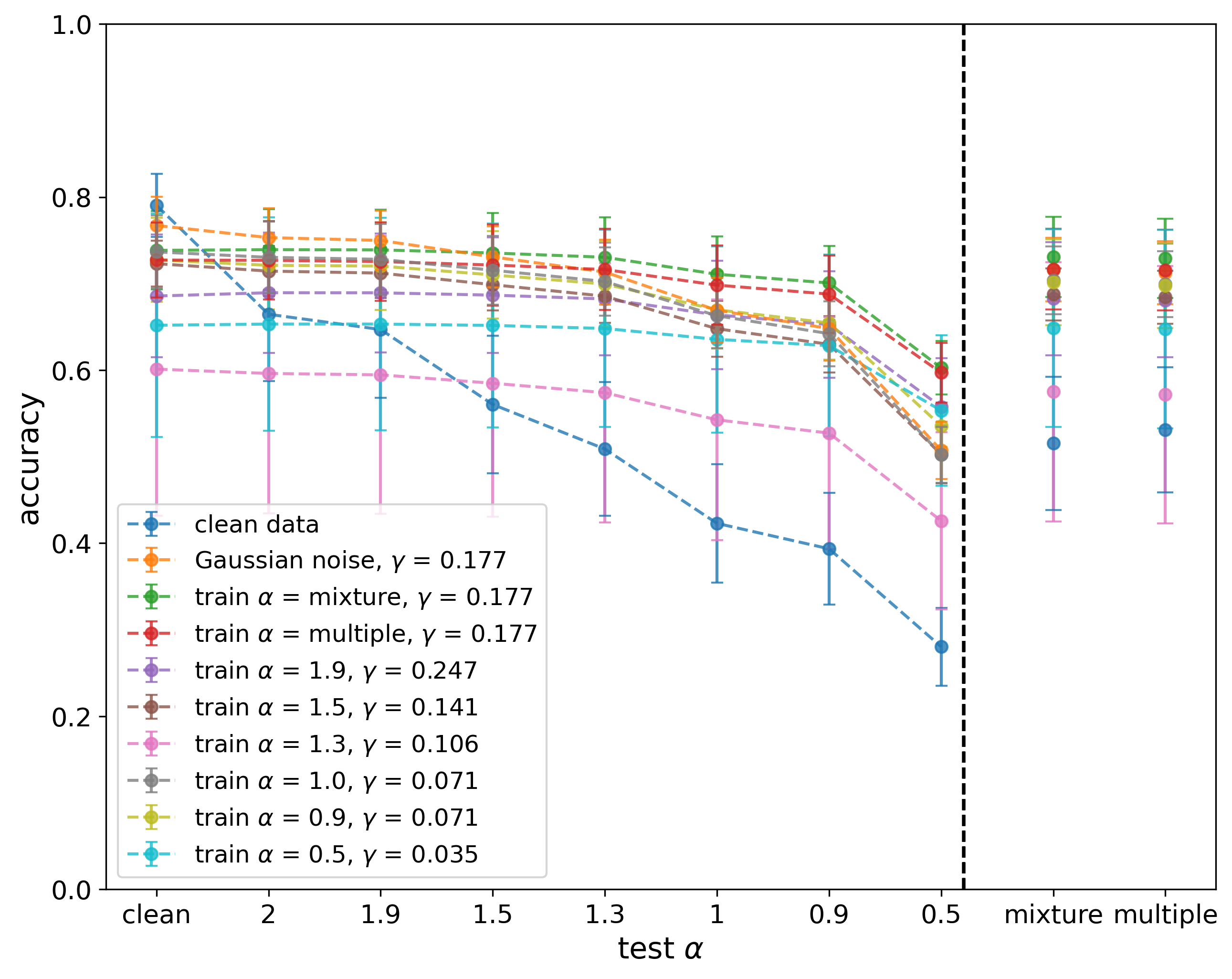}
\caption{width=3, depth=4}
\end{minipage}
\begin{minipage}[t]{0.48\textwidth}
\centering
\includegraphics[width=3in]{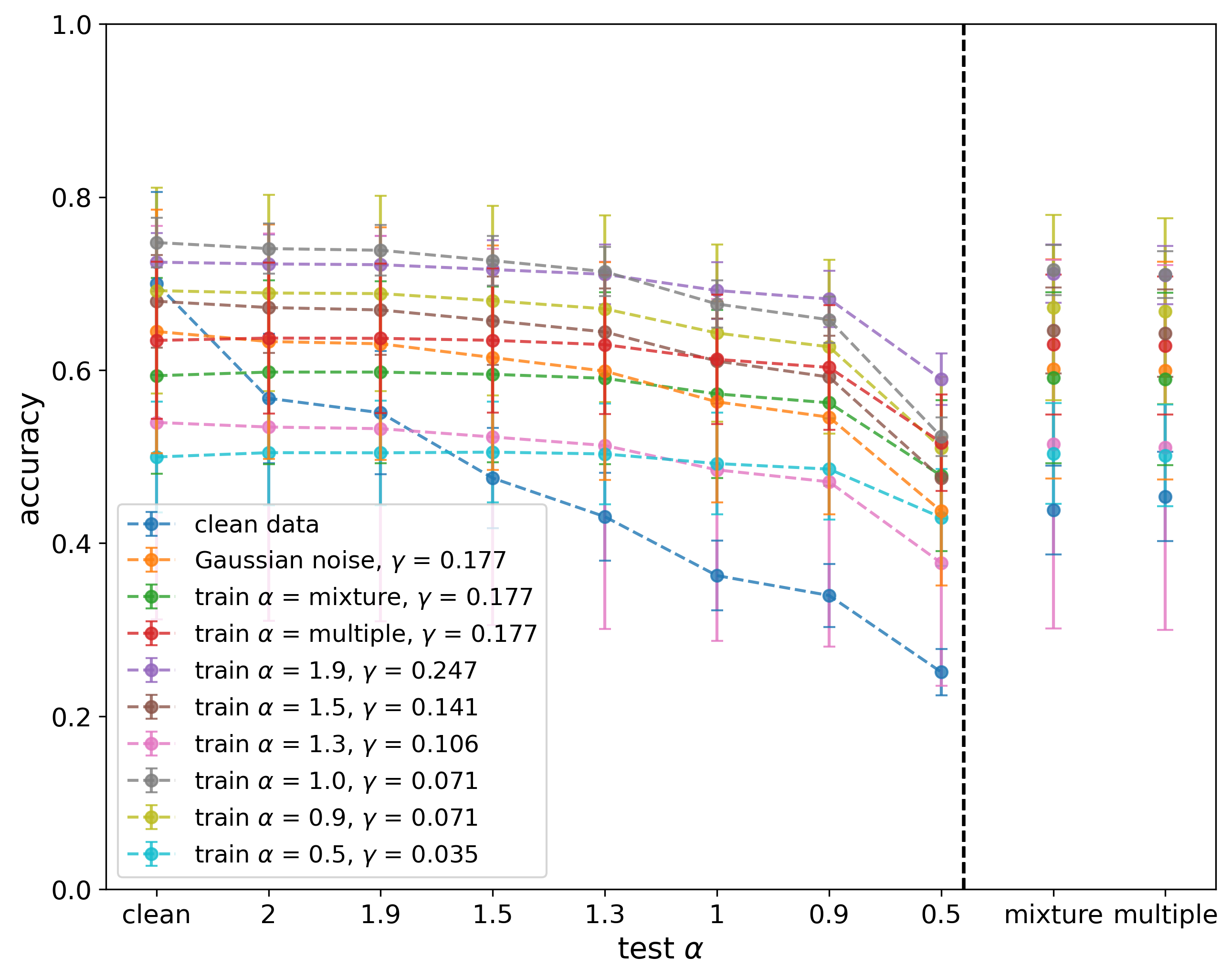}
\caption{width=3, depth=5}
\end{minipage}
\end{figure}

\clearpage

\section{Results of ResNet on CIFAR10}

\begin{figure}[htbp]
\centering
\begin{minipage}[t]{0.48\textwidth}
\centering
\includegraphics[width=3in]{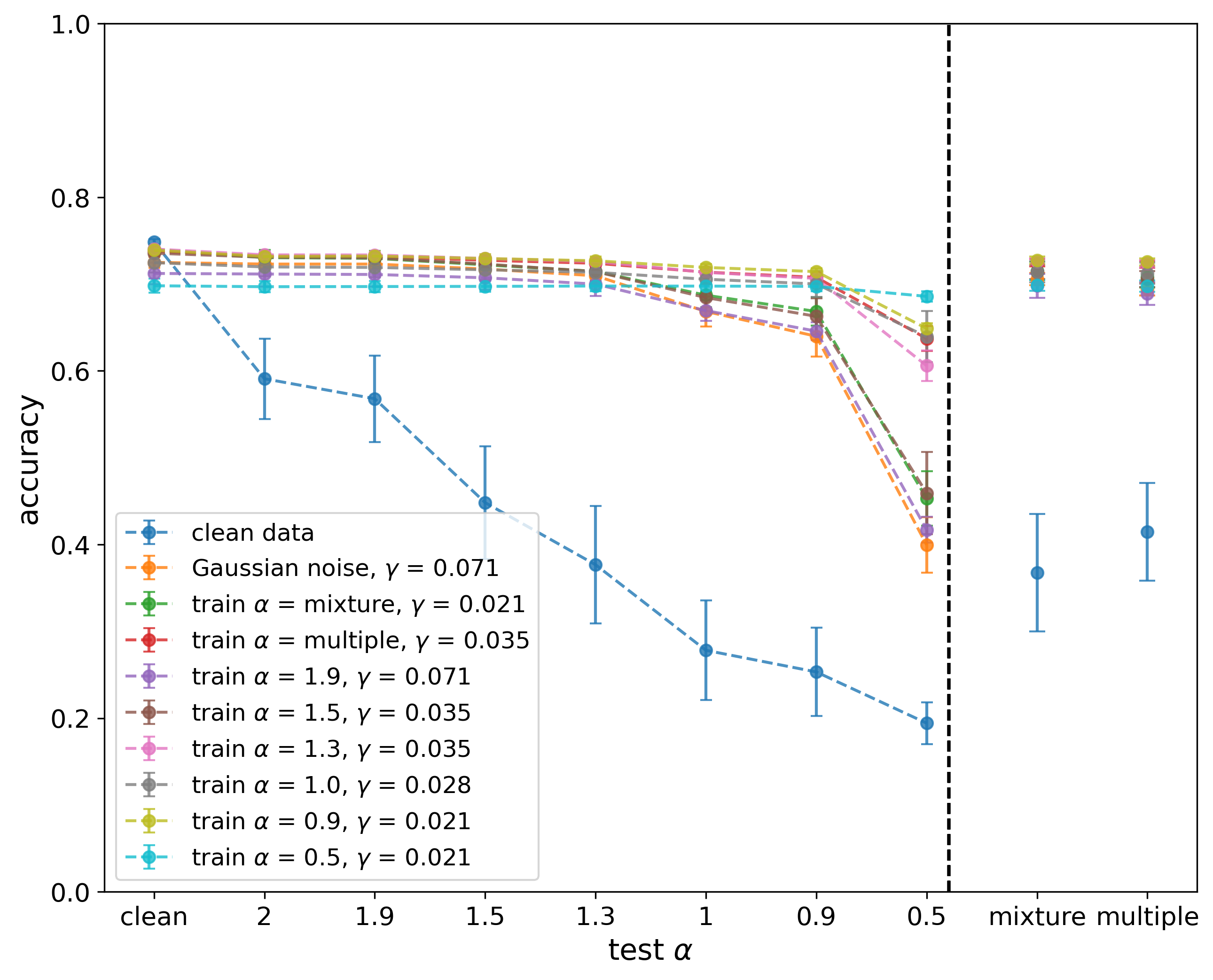}
\caption{width=8, depth=38}
\end{minipage}
\begin{minipage}[t]{0.48\textwidth}
\centering
\includegraphics[width=3in]{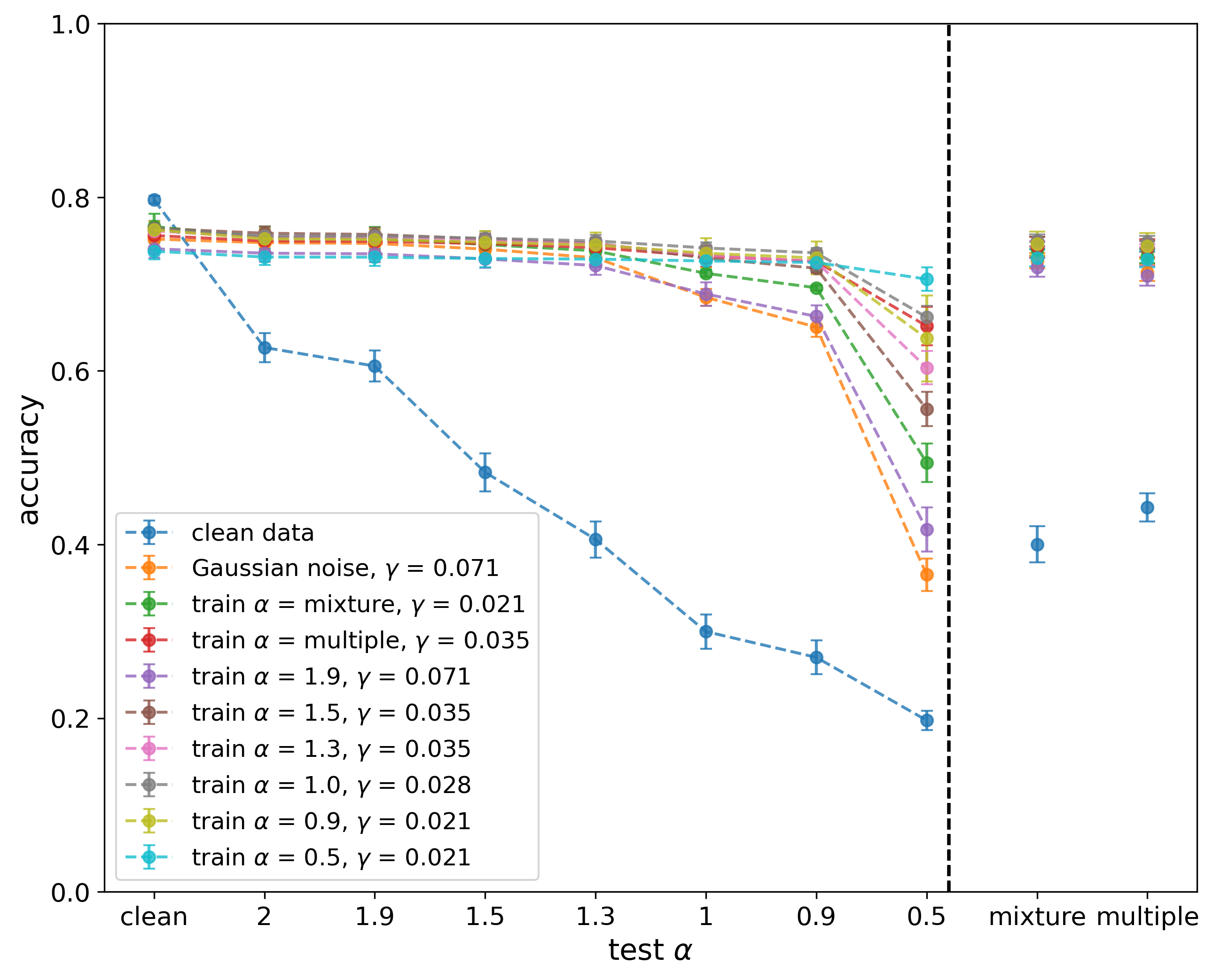}
\caption{width=32, depth=38}
\end{minipage}
\begin{minipage}[t]{0.48\textwidth}
\centering
\includegraphics[width=3in]{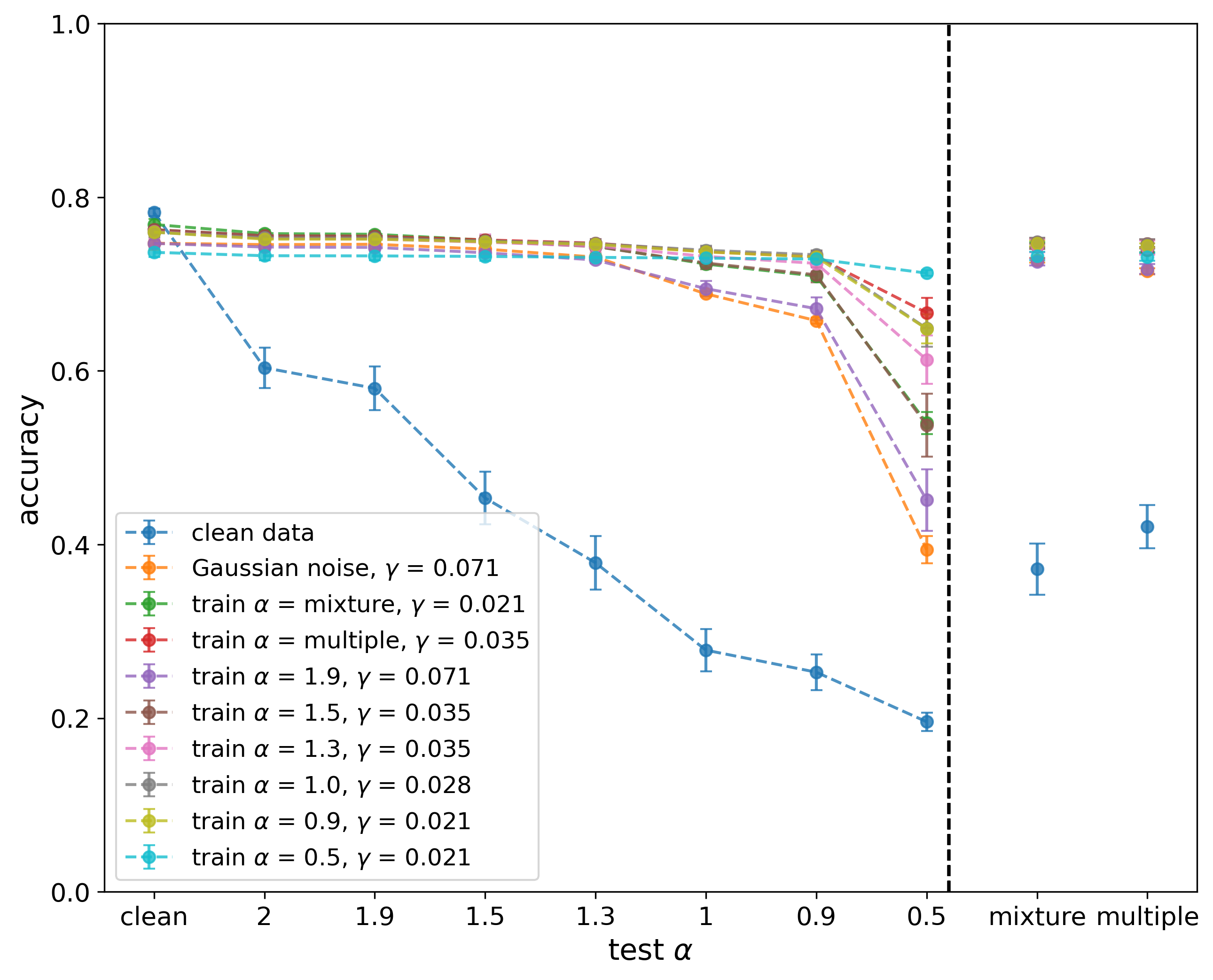}
\caption{width=16, depth=32}
\end{minipage}
\begin{minipage}[t]{0.48\textwidth}
\centering
\includegraphics[width=3in]{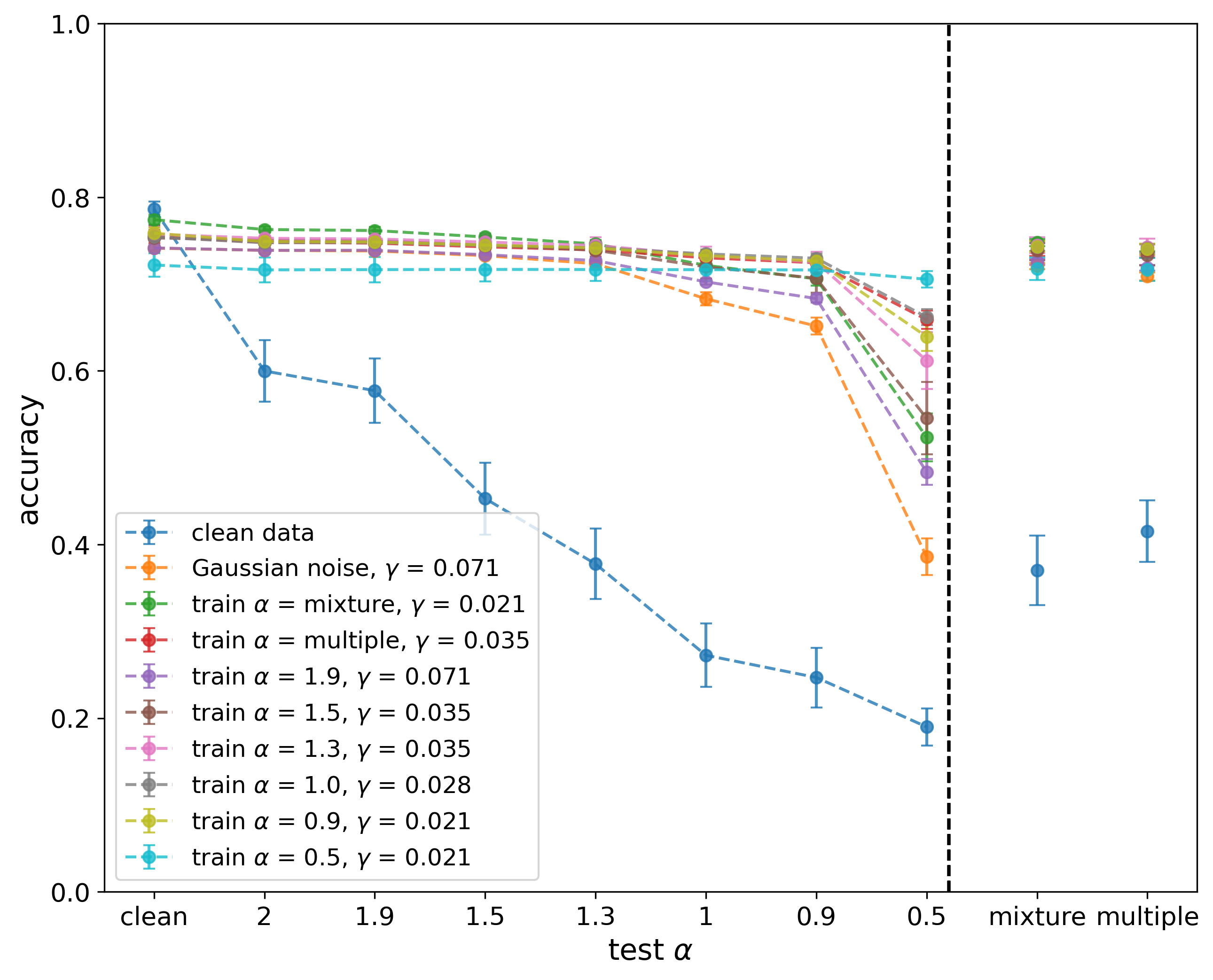}
\caption{width=16, depth=44}
\end{minipage}
\end{figure}
\clearpage
\section{Results of VGG on ECG200}
\begin{figure}[H]
\centering
\begin{minipage}[t]{0.48\textwidth}
\centering
\includegraphics[width=3in]{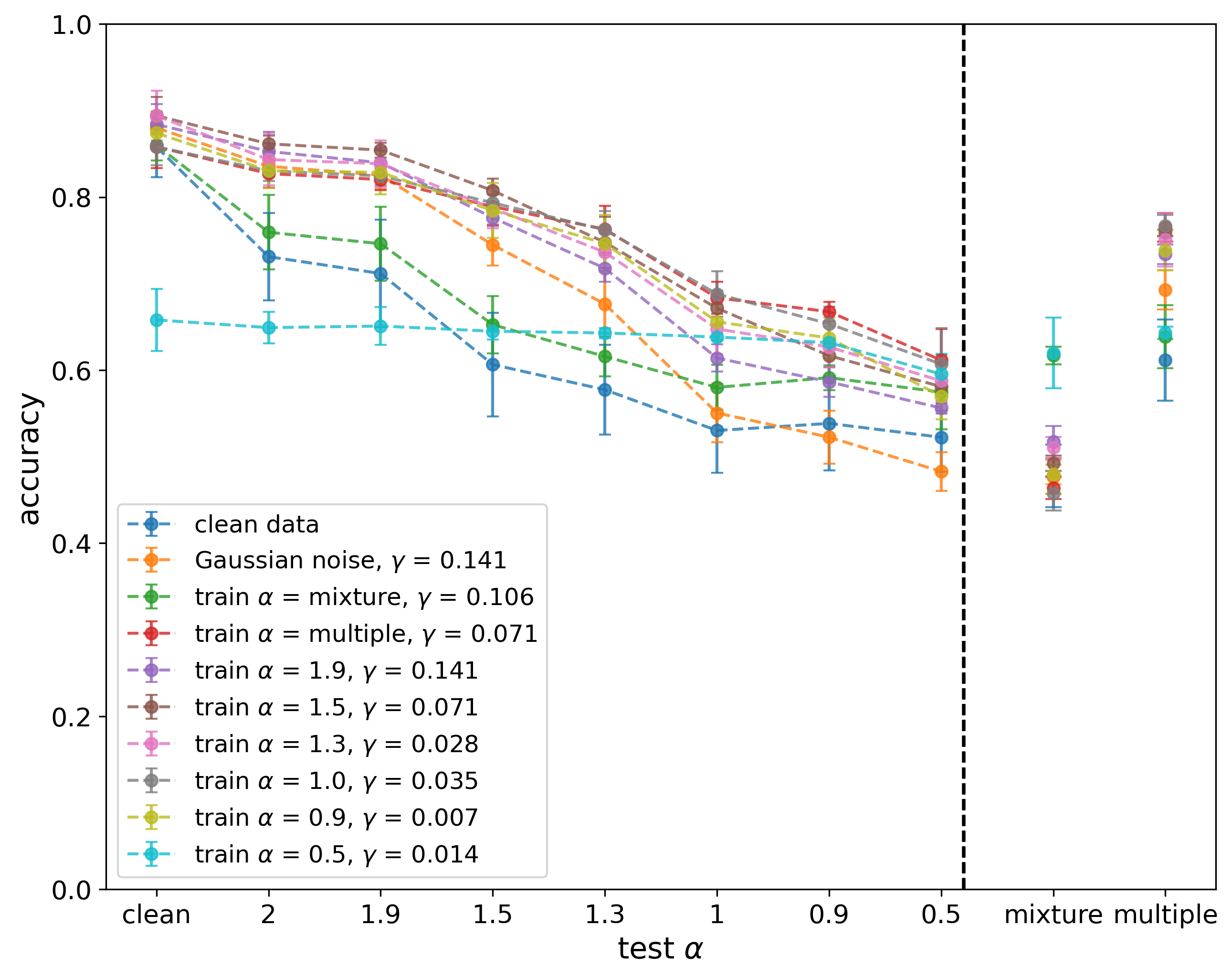}
\caption{width=16, depth=13}
\end{minipage}
\begin{minipage}[t]{0.48\textwidth}
\centering
\includegraphics[width=3in]{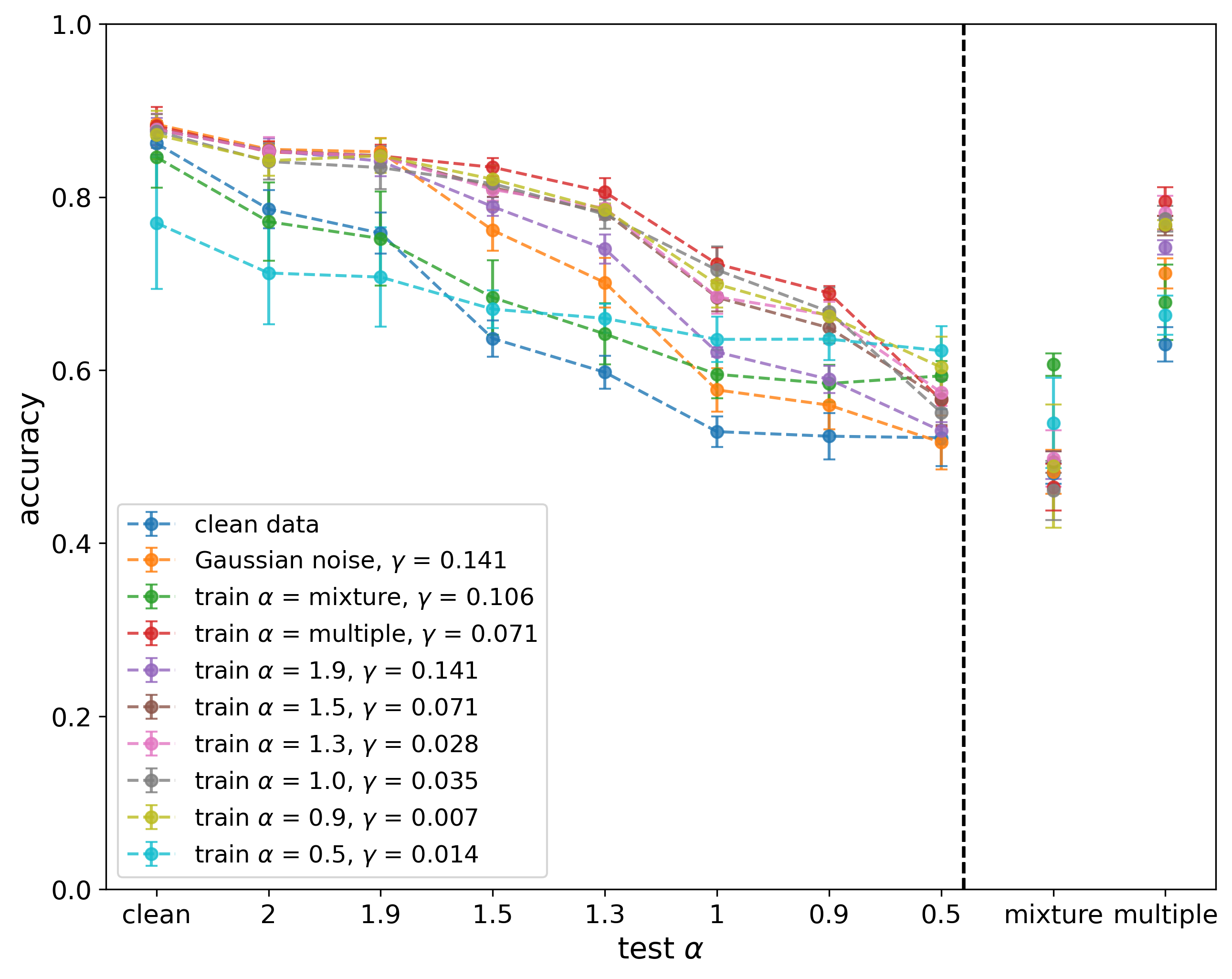}
\caption{width=32, depth=13}
\end{minipage}
\begin{minipage}[t]{0.48\textwidth}
\centering
\includegraphics[width=3in]{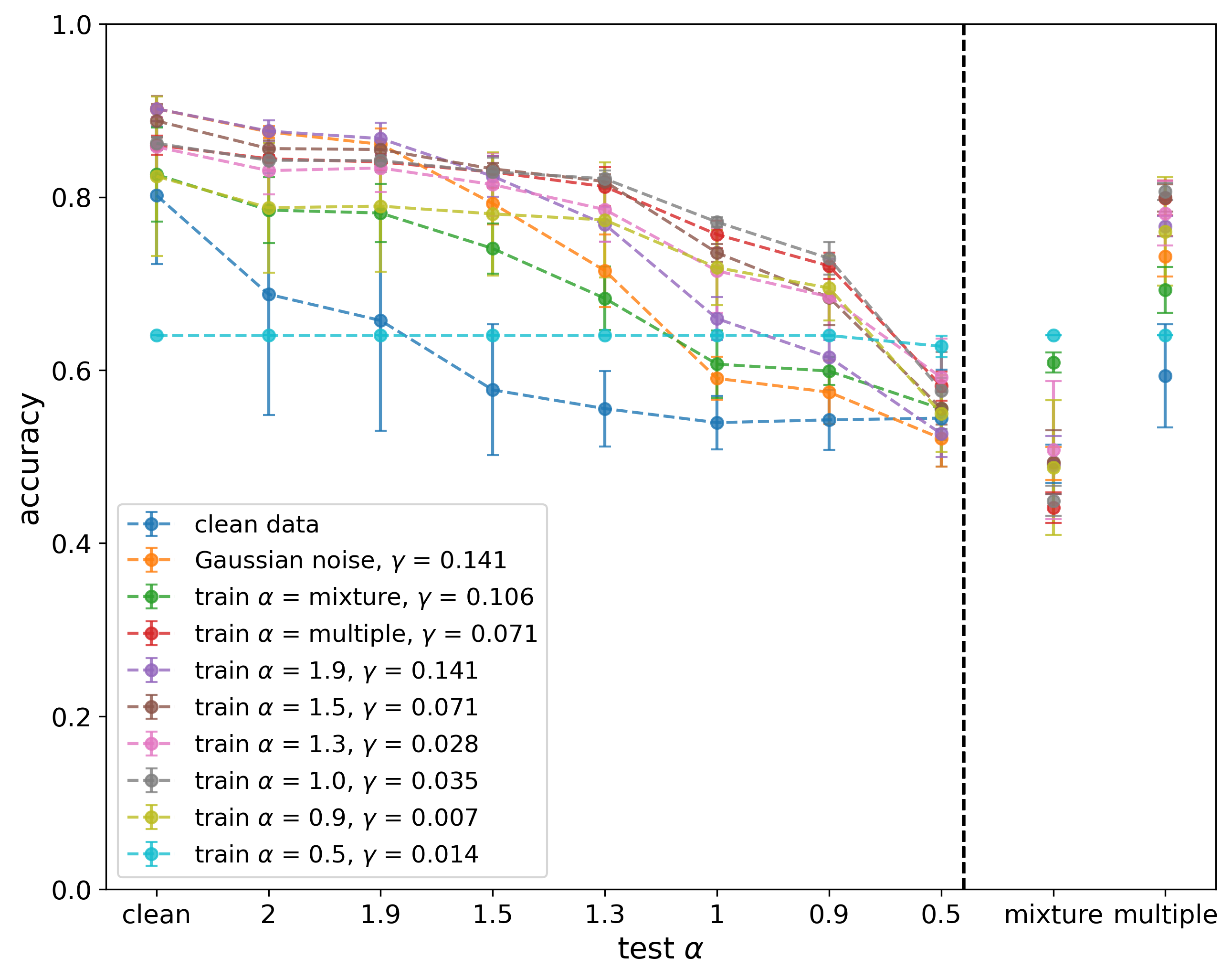}
\caption{width=128, depth=13}
\end{minipage}
\begin{minipage}[t]{0.48\textwidth}
\centering
\includegraphics[width=3in]{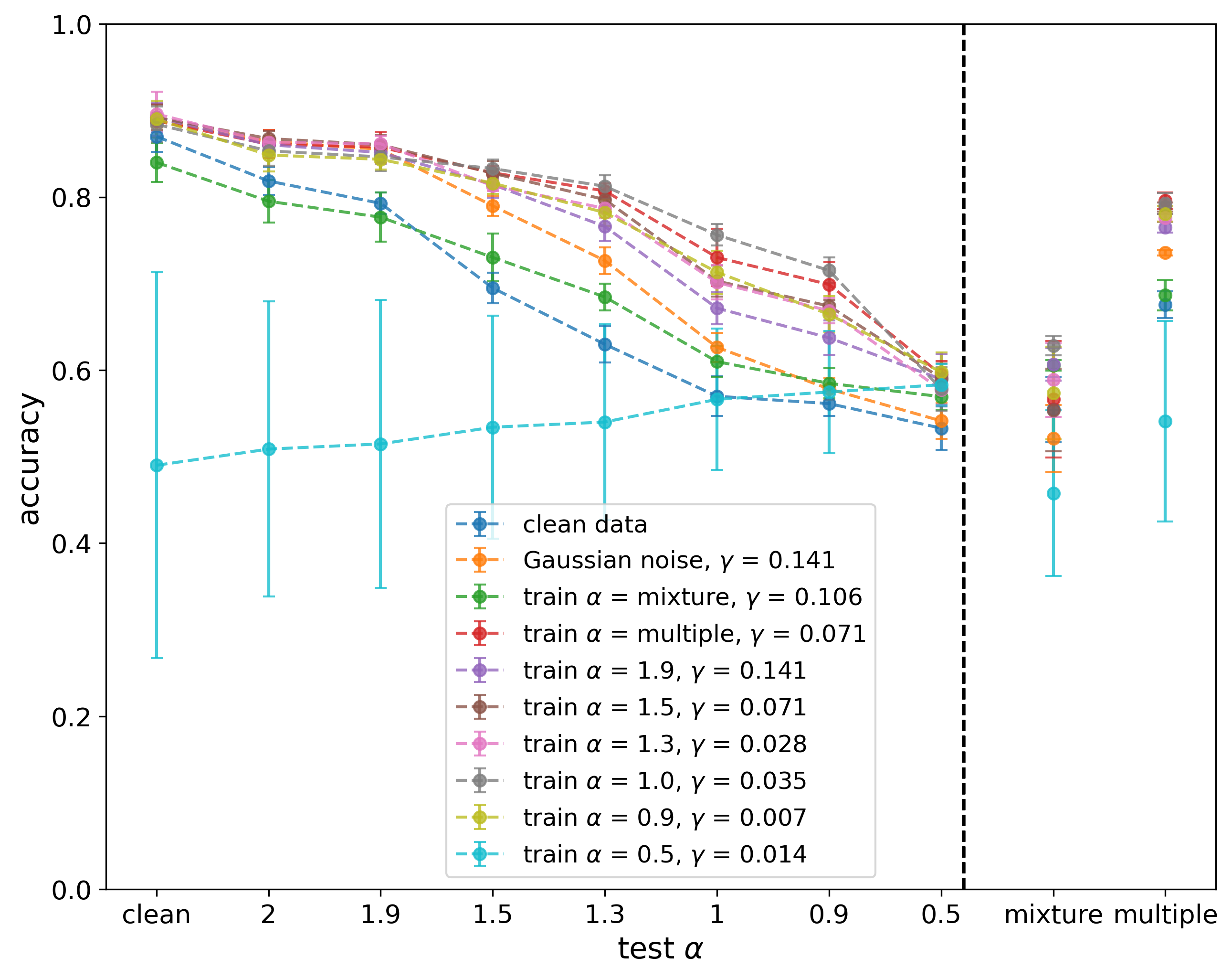}
\caption{width=64, depth=7}
\end{minipage}
\begin{minipage}[t]{0.48\textwidth}
\centering
\includegraphics[width=3in]{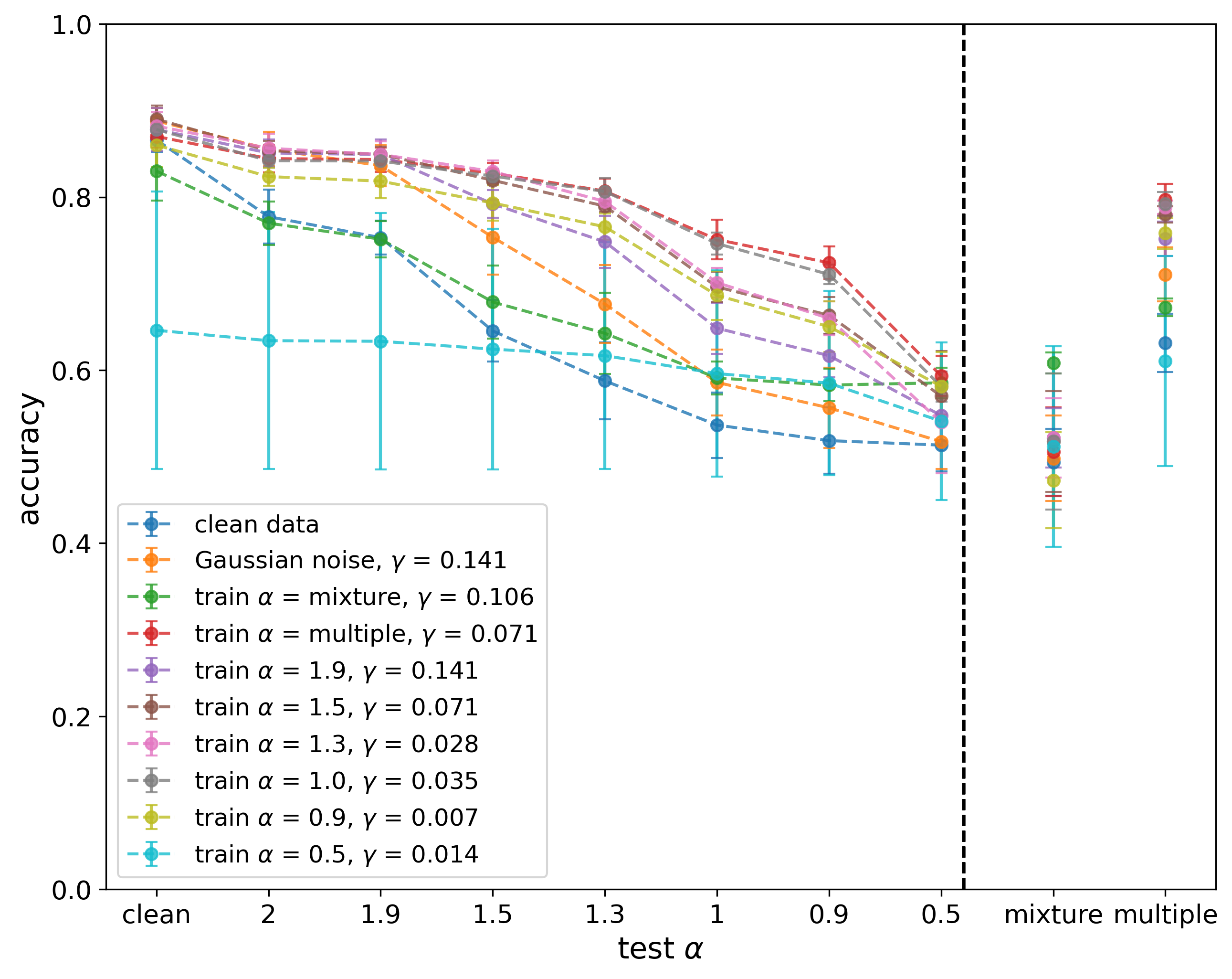}
\caption{width=64, depth=9}
\end{minipage}
\begin{minipage}[t]{0.48\textwidth}
\centering
\includegraphics[width=3in]{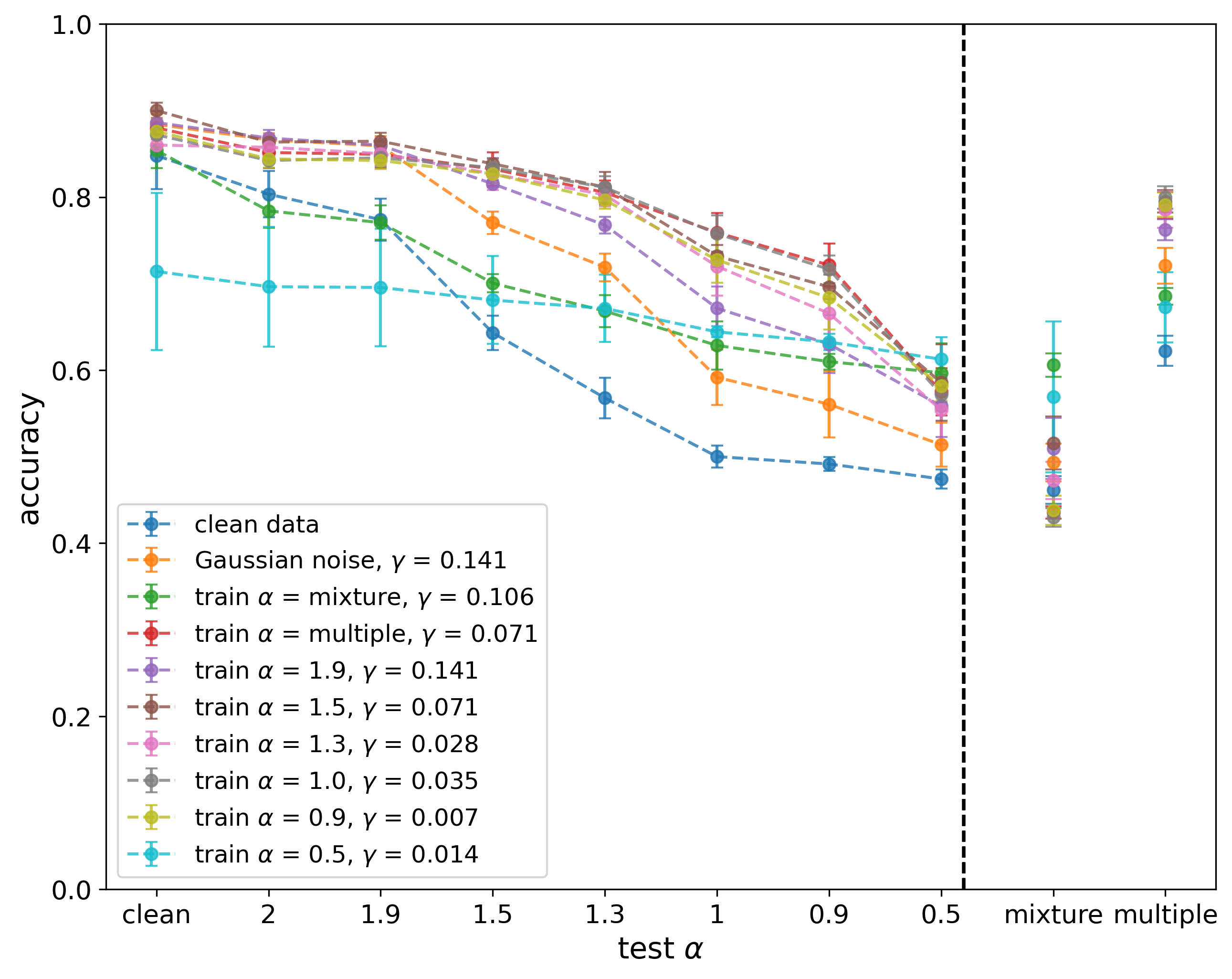}
\caption{width=64, depth=11}
\end{minipage}
\end{figure}

\clearpage
\section{Results of LSTM on LIBRAS}

\begin{figure}[htbp]
\centering
\begin{minipage}[t]{0.48\textwidth}
\centering
\includegraphics[width=3in]{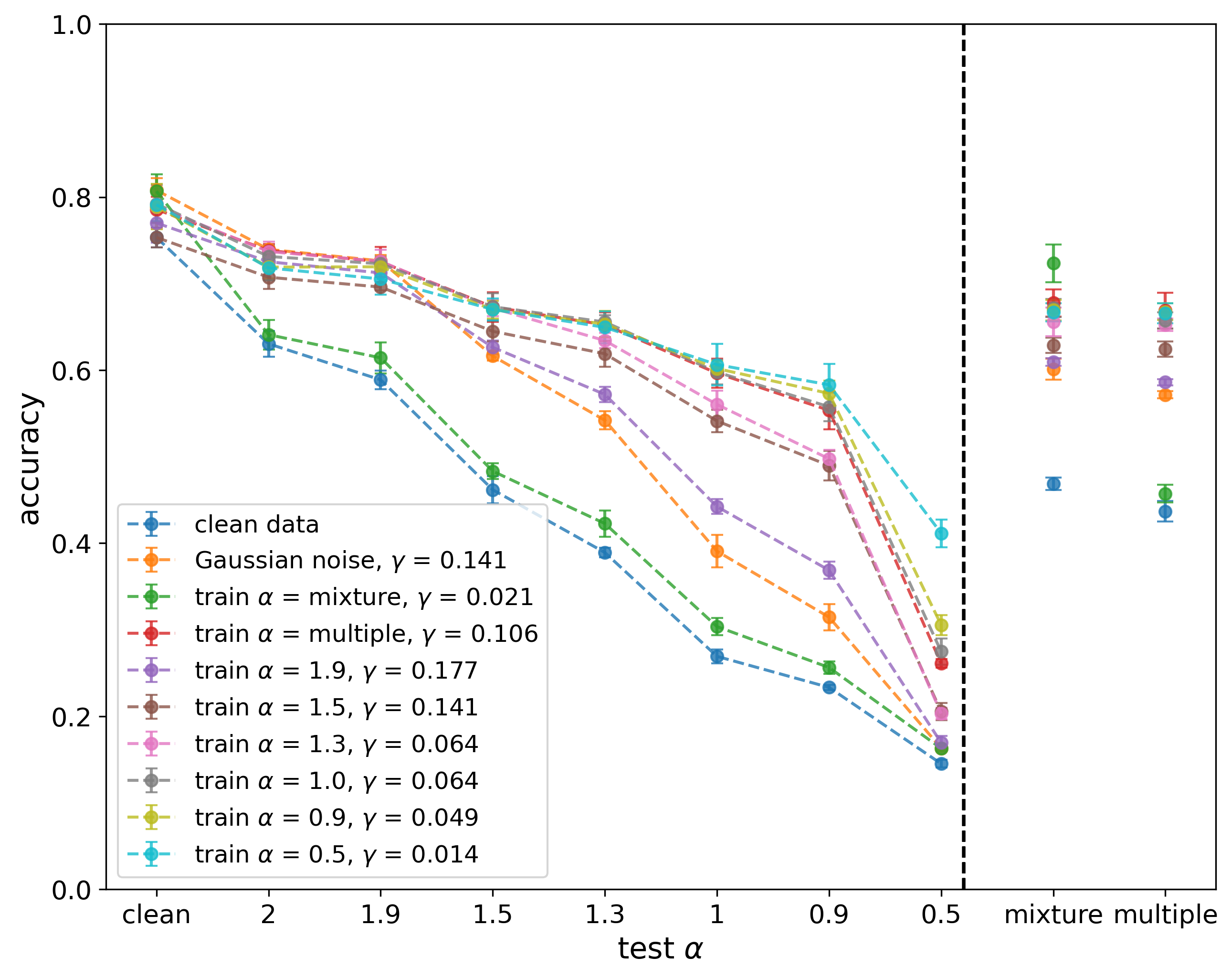}
\caption{width=256, depth=1}
\end{minipage}
\begin{minipage}[t]{0.48\textwidth}
\centering
\includegraphics[width=3in]{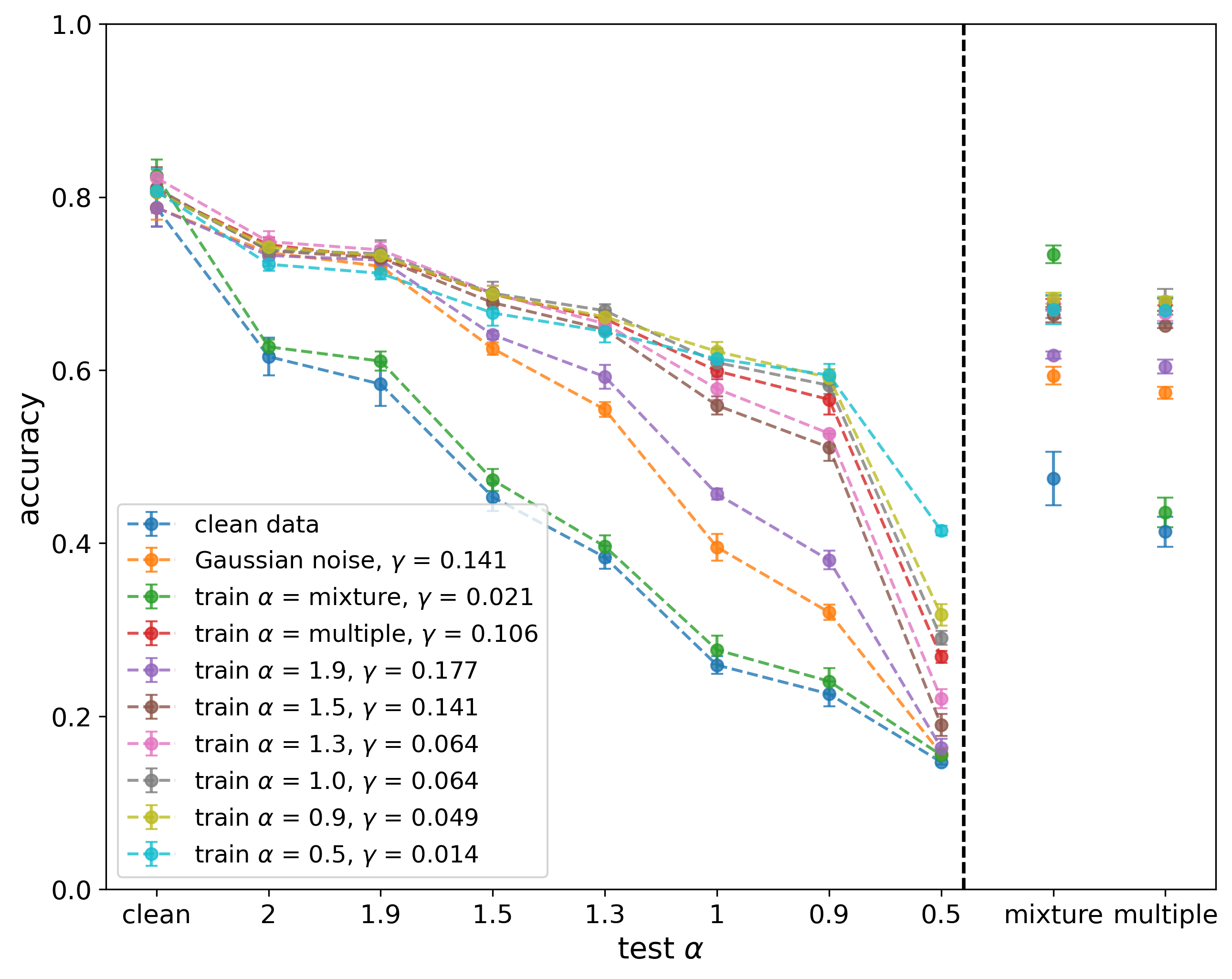}
\caption{width=512, depth=1}
\end{minipage}
\begin{minipage}[t]{0.48\textwidth}
\centering
\includegraphics[width=3in]{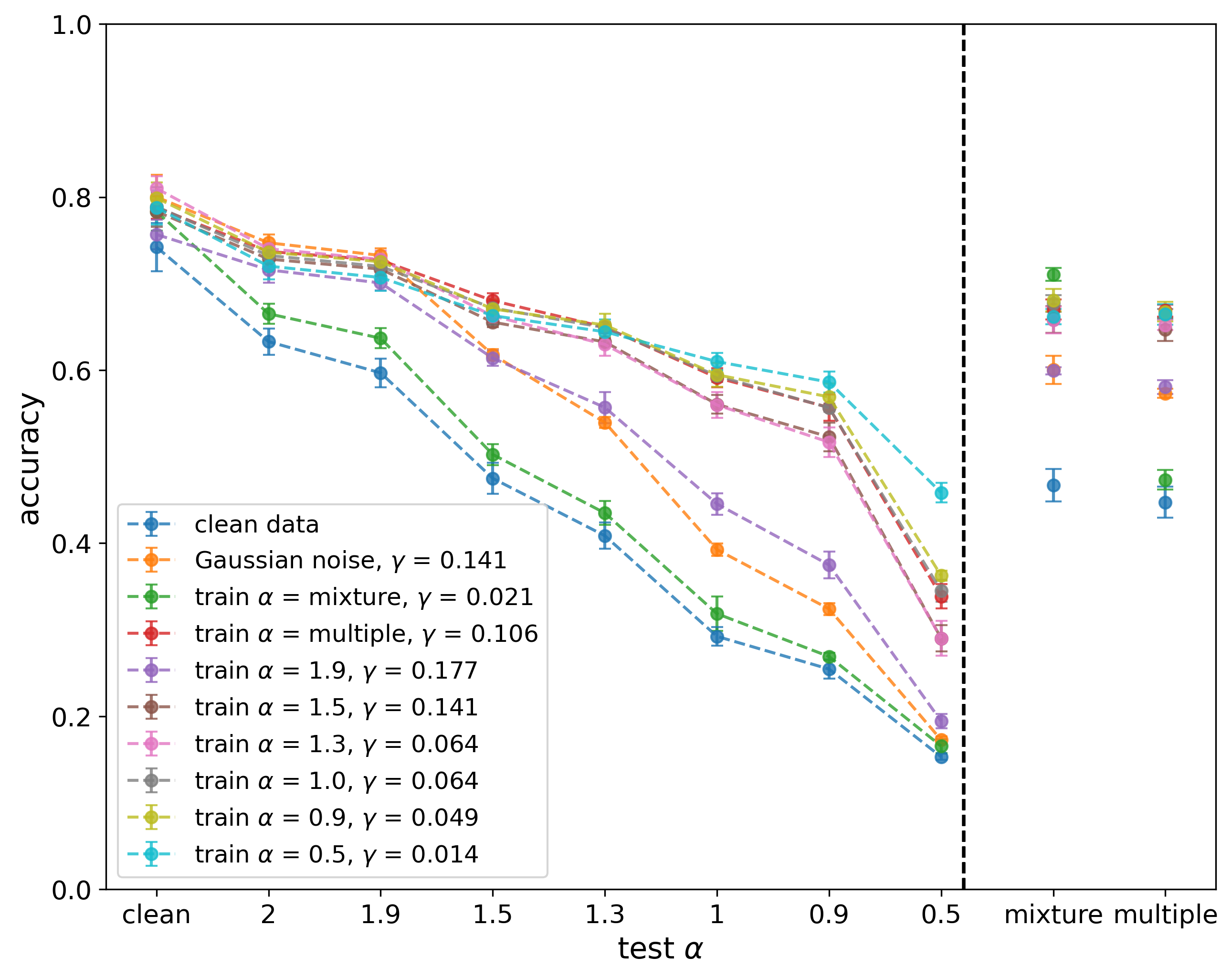}
\caption{width=128, depth=2}
\end{minipage}
\begin{minipage}[t]{0.48\textwidth}
\centering
\includegraphics[width=3in]{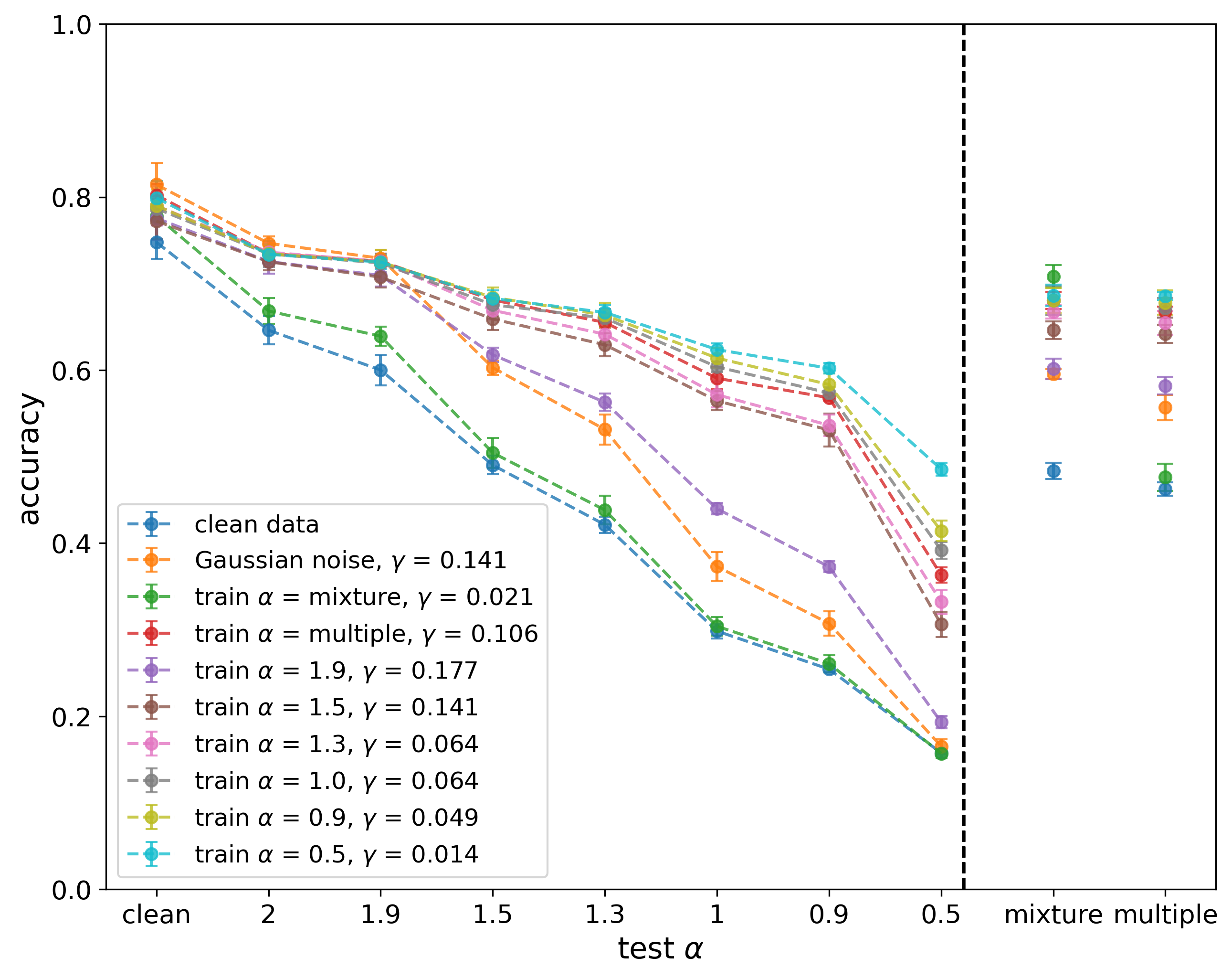}
\caption{width=128, depth=3}
\end{minipage}
\end{figure}

\section*{Checklist}

 \begin{enumerate}

 \item For all models and algorithms presented, check if you include:
 \begin{enumerate}
   \item A clear description of the mathematical setting, assumptions, algorithm, and/or model. [Yes]
   \item An analysis of the properties and complexity (time, space, sample size) of any algorithm. [Not Applicable]
   \item (Optional) Anonymized source code, with specification of all dependencies, including external libraries. [Yes]
 \end{enumerate}

 \item For any theoretical claim, check if you include:
 \begin{enumerate}
   \item Statements of the full set of assumptions of all theoretical results. [Not Applicable]
   \item Complete proofs of all theoretical results. [Not Applicable]
   \item Clear explanations of any assumptions. [Not Applicable]     
 \end{enumerate}

 \item For all figures and tables that present empirical results, check if you include:
 \begin{enumerate}
   \item The code, data, and instructions needed to reproduce the main experimental results (either in the supplemental material or as a URL). [Yes]
   \item All the training details (e.g., data splits, hyperparameters, how they were chosen). [Yes]
         \item A clear definition of the specific measure or statistics and error bars (e.g., with respect to the random seed after running experiments multiple times). [Yes]
         \item A description of the computing infrastructure used. (e.g., type of GPUs, internal cluster, or cloud provider). [Yes]
 \end{enumerate}

 \item If you are using existing assets (e.g., code, data, models) or curating/releasing new assets, check if you include:
 \begin{enumerate}
   \item Citations of the creator If your work uses existing assets. [Yes]
   \item The license information of the assets, if applicable. [Not Applicable]
   \item New assets either in the supplemental material or as a URL, if applicable. [Not Applicable]
   \item Information about consent from data providers/curators. [Not Applicable]
   \item Discussion of sensible content if applicable, e.g., personally identifiable information or offensive content. [Not Applicable]
 \end{enumerate}

 \item If you used crowdsourcing or conducted research with human subjects, check if you include:
 \begin{enumerate}
   \item The full text of instructions given to participants and screenshots. [Not Applicable]
   \item Descriptions of potential participant risks, with links to Institutional Review Board (IRB) approvals if applicable. [Not Applicable]
   \item The estimated hourly wage paid to participants and the total amount spent on participant compensation. [Not Applicable]
 \end{enumerate}

 \end{enumerate}












\vfill